\documentclass[10pt,journal,compsoc]{IEEEtran}

\usepackage{graphicx}
\usepackage{amsmath}
\usepackage{amsfonts}
\usepackage{bm}
\usepackage{makecell}
\usepackage{indentfirst,booktabs,multirow,tabularx}
\usepackage[table]{xcolor}
\usepackage[noend]{algpseudocode}
\usepackage{algorithmicx,algorithm}
\usepackage[colorlinks,bookmarksopen,bookmarksnumbered,citecolor=green, linkcolor=blue, urlcolor=red]{hyperref}
\usepackage{subfigure}

\ifCLASSOPTIONcompsoc
  \usepackage[nocompress]{cite}
\else
  \usepackage{cite}
\fi
\definecolor{mygray}{gray}{.9}

\begin{document}
\title{Surface Geometry Processing: An Efficient Normal-based Detail Representation}

\author{Wuyuan~Xie,~
        Miaohui~Wang$^*$,
        Di~Lin,~
        Boxin~Shi,~
        and~Jianmin~Jiang
		
\IEEEcompsocitemizethanks{
\IEEEcompsocthanksitem Wuyuan Xie, Miaohui Wang, and Jianmin Jiang are with Shenzhen University, Shenzhen, China. 
\IEEEcompsocthanksitem Di Lin is with College of Intelligence and Computing, Tianjin University, Tianjin, China. 
\IEEEcompsocthanksitem Boxin Shi is with School of Computer Science, Peking University, Beijing, China. 
\IEEEcompsocthanksitem The project website of this manuscript can be accessed from https://charwill.github.io/SGP.html, where the implementation and dataset can also be downloaded.
}
\thanks{\textit{Corresponding author}: \textit{Miaohui Wang}}
}

%
%

\markboth{Journal of \LaTeX\ Class Files,~Vol.~0, No.~0, July~2023}%
{Xie \MakeLowercase{\textit{et al.}}: Bare Demo of IEEEtran.cls for Computer Society Journals}
%

\IEEEtitleabstractindextext{%
\begin{abstract}
With the rapid development of high-resolution 3D vision applications, the traditional way of manipulating surface detail requires considerable memory and computing time. To address these problems, we introduce an efficient surface detail processing framework {in 2D normal domain}, which extracts new normal feature representations as the carrier of micro geometry structures that  are illustrated both theoretically and empirically in this article. Compared with the existing state of the arts, we verify and demonstrate that the proposed normal-based representation  has three important properties, including \textit{detail separability}, \textit{detail transferability} and \textit{detail idempotence}. Finally, three new schemes are further designed for geometric surface detail processing applications, including \textit{geometric texture synthesis}, \textit{geometry detail transfer},  and \textit{3D surface super-resolution}. Theoretical analysis and experimental results on the latest benchmark dataset verify the effectiveness and versatility of our normal-based representation, which accepts $30$ times of the input surface vertices but at the same time only takes $6.5\%$ memory cost and $14.0\%$ running time in comparison with existing competing algorithms.
\end{abstract}

}

\maketitle

\IEEEdisplaynontitleabstractindextext

%
\IEEEpeerreviewmaketitle

\IEEEraisesectionheading{\section{Introduction}\label{sec:introduction}}

\IEEEPARstart{S}{urface} geometry detail processing has been studied in mesh domain \cite{cook1984shade, tasdizen2002geometric, tan2018variational, li2020dnf} for decades, where vertices and facets are usually assumed to have an explicit relationship. However, the requirement of pre-meshing limits the existing methods as such that only a simple or virtual surface can be accepted with regular texture patterns. For instance, given a densely scanned surface, it will encounter at least two additional  difficulties: 1) the pre-meshing operation may distort some micro geometry structures, especially for non-regular texture patterns;  2) the pre-meshing operation requires extensive time cost and massive memory consumption. In view of these issues, it is urgent to reconsider and address the surface geometry detail processing from a new perspective, in order to satisfy the growing demand for real and dense 3D surface applications \cite{kumar2019superpixel}.

Traditionally, the definition and operation of geometric details need to be in an ideal mesh environment, where vertices and facets have a clear constraint relationship. The micro geometry structures have been defined as the displacement of a current vertex position relative to the mean value in its neighborhood \cite{wang2003view, wang2004generalized}. Such a definition allows the geometry details to be edited in an intuitive way, but the mesh setting is required to be ideal enough, leading to a tedious preprocessing procedure. In addition, a more comprehensive description for geometry details, namely \textit{voxel} \cite{bhat2004geometric,rematas2020neural}, has been introduced by measuring the geometry pattern on a cubic neighborhood. Although \textit{voxel} can  more accurately describe geometry structure, it costs a huge memory storage. What is more, all these geometric texture definitions are based on a global coordinate system, which is redundant for the local geometry manipulation.

To wisely process the local structures, a relative coordinate system \cite{alexa2003differential,zeng20173dmatch} has been adopted to locally represent the detail features. For example,  the absolute vertex coordinate can be transformed into a relative one via a sparse linear system \cite{alexa2003differential}. The detail feature defined by a relative coordinate has been widely used in mesh editing due to its stability and flexibility. However, these methods are low efficient to deal with dense 3D surfaces, which calls for a bunch of adjacent information during the absolute-relative coordinate transformation.
\begin{figure*}[!t]
\centering 
\includegraphics[width=0.98\linewidth, ]{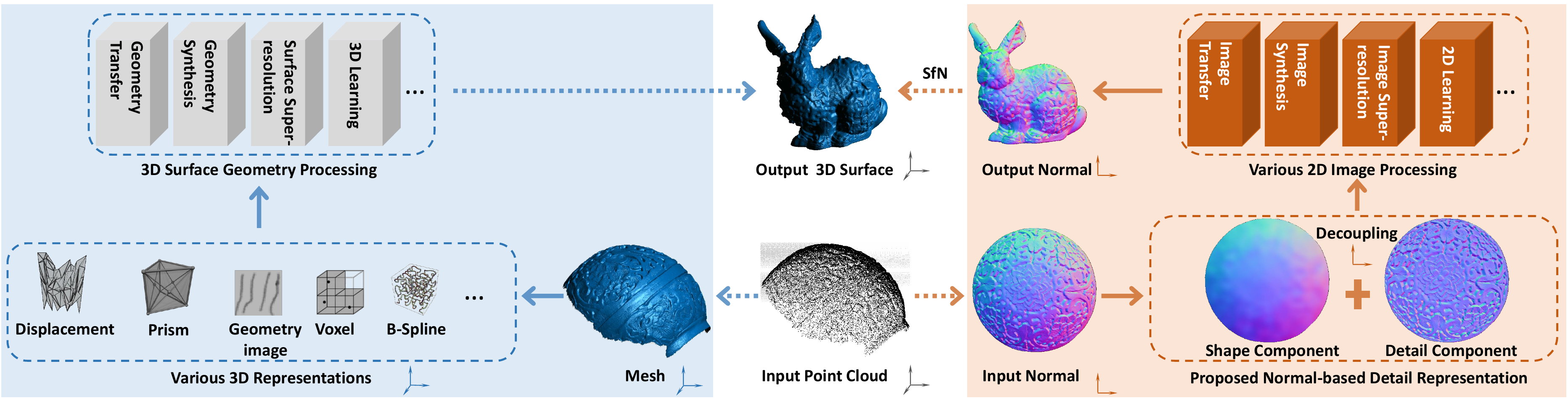}
\caption{\textbf{Illustration of the difference of surface geometry detail processing between the traditional 3D representations and the proposed 2D normal-based detail representation.} Our method manipulates the 3D surface details in 2D normal domain, which aims to decouple a surface normal map into a \textit{shape component} and \textit{detail component}. We theoretically derive that the proposed \textit{detail component} has the properties of separability,  transferability, and idempotence.} 
\label{fig_overall_illu} 
\end{figure*}

Recently, deep learning has achieved a remarkable success in various computer vision tasks \cite{wang2021learning,chen2020deep,ju2021recovering}, which has also been employed for geometry detail processing. For instance, for \textit{surface detail transfer}, the mapping mechanism  between geometric texture and 3D shape can be trained in an end-to-end manner by  convolutional neural network (CNN) \cite{berkiten2017learning}. For \textit{geometry texture synthesis}, metric learning can synthesize the small fragment of a given geometry texture on the entire target surface \cite{hertz2020deep}. However, there is  still a large room for  performance improvement on learning-based geometry detail processing. On  one hand, an internal network architecture is usually difficult to be clearly interpretable, and thus the robustness and generalization of learned models cannot be guaranteed for new data samples. On the other,  a deep-learned model is usually complicated, and requires a high computation  cost.

To overcome the above limitations, the normal vector of a surface shape has been used as the carrier of geometric details. According to the photometry principle \cite{bass2009handbook}, the diversity of a surface normal vector contains fine-grained differences in terms of illumination intensities, which is perceived as texture (or geometry details) by human eyes. Besides, the normal map of a 3D surface also conveys a  global shape geometry, as demonstrated in \cite{tasdizen2003geometric,xie2014surface}. Therefore, a surface normal map encodes both the global shape and the local geometry structure simultaneously.

Inspired by the above discussions, we present a new simple and  interpretable framework for geometric detail processing in normal domain. One of the key advantages is that the heavy payload of meshing can he avoided, and the legacy image processing algorithms can be re-used. Unlike those existing approaches, our method rarely relies on the mesh environment.  For instance, it can accept real and dense normal data  obtained from a real-world surface by \textit{photometric stereo} \cite{ikehata2014photometric,xie2015photometric,cho2018semi}. More importantly, we excavate an effective geometry detail representation which allows to be easily processed as a digital image. This is of significant importance for many popular applications in 3D scenarios, due to the fact that most of them have dense and complex micro geometry structures. In addition, surface in the mesh format can be considered as a special case of the proposed framework by recording the patch orientation vectors into a dense normal map which can be easily achieved via \textit{MeshLab} (see a dense laser-scanning point cloud data in Section \ref{sec:experiments}). 
 The proposed normal-based detail representation is highly effective for surface geometry processing (\textit{e.g.}, \textit{geometric texture synthesis}, \textit{geometry detail transfer},  \textit{3D surface super-resolution}, \textit{etc}.), which can be implemented in a light-weight image processing manner as illustrated in Fig. \ref{fig_overall_illu}.

We highlight our contributions as follows:
\begin{itemize}
\item Based on the characteristics of a surface normal representation, we propose to manipulate surface geometry details by excavating the corresponding surface normal into two descriptors: \textit{surface shape} and \textit{geometry detail}. As far as we know, this is an earlier exploration to comprehensively manipulate 3D surface detail in a light-weight image processing manner.

\item To better analyze the \textit{geometry detail} descriptor,  we further derive and demonstrate three important properties, including \textit{detail separability}, \textit{detail transferability}, and \textit{detail  idempotence}. These properties mathematically guarantee that {the \textit{geometry detail} of a source surface can be used as a  standalone feature applied to a target shape}.

\item Based on the proposed normal-based representations, we further design three new schemes for geometric surface detail processing, including \textit{geometric texture synthesis}, \textit{geometry detail  transfer}, and \textit{3D surface super-resolution}. In addition, recent deep-learned models can be seamlessly embedded into the proposed framework. Experiments on the benchmark dataset verify the effectiveness and versatility of our approach over recent competing schemes in terms of both the computing time and memory spaces.
\end{itemize}

The remainder of this article is organized as follows. Section \ref{sec:relatework} gives a concise background introduction for recent advances in surface geometry processing.
Section  \ref{sec:georepresentation} presents the geometric surface representation in the normal domain. 
In Section \ref{sec:detailproperty}, we describe three main properties of the proposed geometric processing framework, including \textit{detail separability}, \textit{detail transferability}, and \textit{detail  idempotence}.  
Section \ref{sec:experiments} demonstrates the significant theoretical and practical contributions of this normal-based surface detail representation, and Section \ref{sec:discussions} provides a discussion for the characteristics of  the proposed scheme.  
Finally, the overall conclusion is drawn in Section \ref{sec:conclusion}, and some open ideas are presented to address several remaining challenges for our normal-based geometry processing.

\section{Background}
\label{sec:relatework}
For simplicity, we briefly introduce the recent developments of geometry detail processing schemes based on the following four questionable aspects: 1) which domain is the texture pattern defined? 2) whether does it require parameterization? 3) how is the capability in dealing with dense data (measured in patch or pixel)? 4) how is the computational complexity (measured in  memory or time)? In general, the first three factors determine the last one. In Table \ref{tab:methods_intros}, we summarize an overall comparison for recent representative methods in surface geometry detail processing.

The proposed method takes a normal map as the input, indicating that it can accept a high-resolution two-dimensional data format. In other words, our method can process a highly-dense geometry texture as a digital image. In addition, the input normal can contain real-world and non-repetitive surface textures, and thus the proposed method is an effective detail representation of surface geometry. Moreover, the normal manipulation does not require a parameterization or mesh environment.

\noindent \textbf{Vertex Displacement-based}. 
The most straightforward way to process geometry details is to represent the texture pattern as a vertex displacement between the target shape and  the original surface \cite{cook1984shade}. This kind of method is too strict and requires more accurate alignment between the target and the original surface \cite{kobbelt1998interactive, ebert2003texturing}. As a result, mesh refinement has to be added to improve the vertex displacement by  providing finer texture patterns, examples of which include multi-resolution mesh \cite{guskov1999multiresolution}, Laplacian mesh \cite{sorkine2004laplacian}, or statistical mesh  \cite{golovinskiy2006statistical}. However, the mesh refinement heavily depends on the parameterization, which degrades its applicability. 
\begin{table}[!t]
\centering
\caption{Comparisons of the representative schemes for 3D surface geometry detail processing.} 
\newsavebox{\tablebox}
\begin{lrbox}{\tablebox}
\setlength{\tabcolsep}{3.mm}
\renewcommand\arraystretch{1.2}
\begin{tabular}{llccr}
\hline 
\hline 
Method	&Domain	&Parameterization	&Density	&Complexity\\
\hline 
\cite{cook1984shade}	&Mesh	&Yes	&Patch/pixel	&Medium\\
\rowcolor{mygray}\cite{kobbelt1998interactive}	&Mesh	&Yes	&Patch/pixel	&Medium\\
\cite{guskov1999multiresolution}	&Mesh	&Yes	&Patch	&High\\

\rowcolor{mygray}\cite{biermann2002cut}	&Mesh	&Yes	&Patch	&Medium\\
\cite{sorkine2004laplacian}	&Mesh	&Yes	&Patch	&High\\ 

\rowcolor{mygray}\cite{finkelstein1994multiresolution}	&B-spline	&Yes	&Patch	&High\\
\cite{praun2001consistent}	&Wavelet	&Yes	&Patch	&Medium\\
\rowcolor{mygray}\cite{botsch2003multiresolution}	&Prism 	&Yes	&Patch	&High\\
\cite{bhat2004geometric}	&Voxel	&Yes	&Patch	&High\\
\rowcolor{mygray}\cite{lai2005geometric}	&Geometry image	&Yes	&Patch/pixel	&Medium\\

\cite{berkiten2017learning}	&Mesh	&No	&Patch	&Medium\\
\rowcolor{mygray} \cite{hertz2020deep}	&Mesh	&No	&Patch	&Medium\\

\hline
\textbf{Proposed}	&Normal	&No	&Patch/pixel	&Low\\
\hline 
\hline 
\end{tabular} 
\end{lrbox}
\scalebox{0.75}{\usebox{\tablebox}}	
\label{tab:methods_intros}
\end{table}

\noindent \textbf{Domain-guided}.  
It was once popular to transfer the texture information to another domain for effective processing. Due to the implicit expression in texture description, B-spline curve \cite{finkelstein1994multiresolution} has been employed to fit both the source and target surface, but the parameter-tuning of the B-spline curve for a complex surface is another major challenge.
In the earliest domain-guided approaches, the source surface and target shape have been  also  transferred to the frequency domain \cite{praun2001consistent} to separate the details and shape as a high-frequency and low-frequency coefficient, respectively. 
Recently, some new representations have been used to process surface details, such as displacement volume \cite{botsch2003multiresolution}, voxel-based \cite{bhat2004geometric},  and geometry image \cite{lai2005geometric}. Although some complex and non-repeating texture patterns can be accurately addressed, the massive computation for auxiliary vertexes or surface re-meshing makes them difficult to be extended in dealing with dense surfaces.

\noindent \textbf{Data-driven}. 
Surface editing of the source texture to the target shape can be considered as a mapping process. Data-driven approaches \cite{ berkiten2017learning, hertz2020deep} have recently been leveraged to directly learn the mapping between the source and the target, as such that a  geometry texture pattern can be automatically synthesized and transferred. This kind of method can effectively avoid the parameterization and the definition of textures. However, at present, they are difficult to be interpretable, and require very high computing resources.

Undoubtedly, the data-driven framework will be one of the main streams  to solve 3D problems in future, but currently it is not universal, especially when high-dimensional geometric features are involved. If a  geometry detail representation can be dimensionally reduced to the 2D  pixel domain, various well developed image-based learning networks can be leveraged to solve existing 3D surface processing problems. This intuitive idea directly inspires this work.

\section{Geometric Surface Representation in Normal Domain}
\label{sec:georepresentation}
\noindent \textbf{Normal Map Definition}.  
A normal map  $\mathcal{N} \in \mathbb{R}^{3\times H \times W}$ (\textit{e.g.}, 3 channels,  height $H$, and width $W$) is such an image that can be obtained by virtual or real imaging projection of a target surface under a single view. The normal map $\mathcal{N}$ has three channels, where each normal pixel represents a normalized normal vector (\textit{i.e.}, xyz components) on its corresponding surface point. 
Now let us backproject $\mathcal{N}$ onto its 3D surface and consider it in the 3D discrete settings. Specifically, for a 3D surface, it can be regarded as composed of numerous small patches (or facets) \cite{wei2020selective,kuo2021surface}. Globally, every patch has a specific orientation, and is connected with each other to form a 3D shape. At the same time, the orientation difference among adjacent patches causes a local surface roughness, \textit{i.e.}, micro geometric structures. The orientation of each surface patch is projected onto the viewing plane to obtain the corresponding normal vector. Thus, a normal map simultaneously encodes both the global shape and the local geometry detail (or micro structure/texture) of its projected surface.

\noindent \textbf{Normal Map Generation}. 
For \textit{photometric stereo}, the normal of a 3D surface point corresponding to the normal pixel can be calculated from the pixel magnitude of multiple images according to the principle of \textit{Lambertian} reflection \cite{xie2015photometric,cho2018semi}. For a \textit{3D mesh} or \textit{point clouds}, the normal map can be generated through orthographic or perspective  projection of a virtual camera. It is noted that our method is independent of normal map generation from orthographic and perspective projections. For a normal map generated from \textit{photometric stereo}, the practical imaging process is under the perspective projection. For a normal map generated from point cloud data, we choose the orthographic projection to facilitate the calculation process. 
Specifically, at a fixed view angle,  visible surface points will be projected onto a virtual image plane, and then their normal vectors will be recorded on the corresponding pixel coordinates to form a normal map, where the normal vector is a fitted local surface orientation of all points in a small region. A detailed description of normal map estimation for point clouds is referred to here\footnote{https://pcl.readthedocs.io/projects/tutorials/en/latest/normal{\_}es\\timation.html?highlight=normal{\%}20estimation}. Therefore, the generation of normal maps is as convenient as taking photographs, but they have the same constraints of single-view imaging.

Based on the observation that, by increasing the diversity of normal orientations while keeping their original low-frequency components,  the geometry surface details can be enriched without distorting its global shape, and we come up with a new extraction method for geometry details from a surface normal map, and show its high transferability to another surface with  different shapes. In the rest of this section, we show the possibility of excavating  these two feature representations (\textit{i.e.}, global shape and local geometry details) from a single normal map, and give their mathematical expressions.

\begin{figure}[!t]
\centering 
\includegraphics[width=0.40\textwidth]{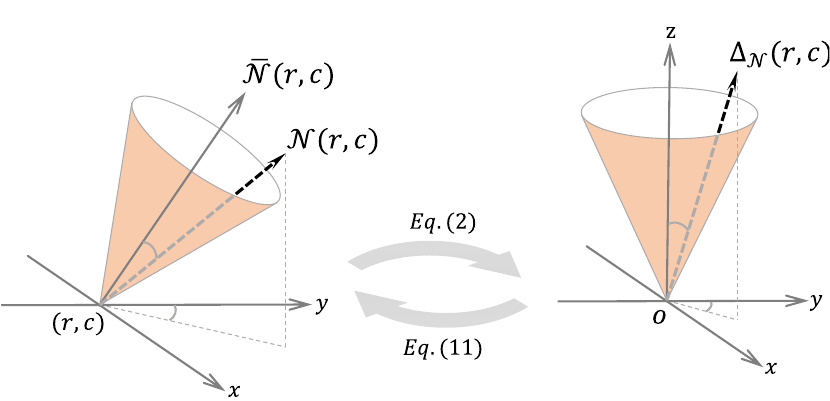}
\caption{\textbf{Illustration of the relative coordinate transformation for extracting the \textit{detail component} representation in Section \ref{sub:detailDef}.} } 
\label{fig_relativeCoord} 
\end{figure}

Specifically, a given normal map $\mathcal{N}$ is used to obtain a \textit{shape component} and a \textit{detail component}, where the \textit{shape component} and the \textit{detail component} are calculated by Eq. (\ref{eq_conv_cal}) in Section \ref{subsec:shapeComponent} and Eq. (\ref{eq_detail_form}) in Section \ref{sub:detailDef}, respectively. 
Geometrically, the relationship between $\mathcal{N}$ and its \textit{shape component} and the \textit{detail component} can be also illustrated as the transformation process between the global coordinate system (left) and the local coordinate system (right), as shown in Fig. \ref{fig_relativeCoord}. 

\subsection{Surface Shape Representation}\label{subsec:shapeComponent}
After removing the fine-grained texture structure from an object surface, {the result of  smoothing} can be {considered to only convey shape information} \cite{chern2018shape,li2021sp}. To derive the surface smoothing operation in the normal domain, we observe that a surface roughness decreases as the variance of the local neighborhood normals decreases. Consequently, if there is a way to reduce the local variance,  a global surface shape can be obtained based on {its normal map}. In view of this, we adopt the smoothing operation as a low-pass filter on the normal map $\mathcal{N}$.	
\begin{figure*}[!t]
\centering
\begin{tabular}{p{3cm}<{\centering} p{3cm}<{\centering} p{3cm}<{\centering} p{3cm}<{\centering} p{3cm}<{\centering}}
		    \includegraphics[height=3.3cm]{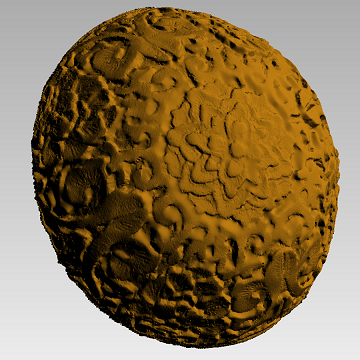}
			&\includegraphics[height=3.3cm]{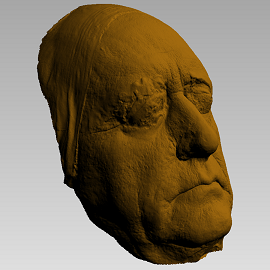}
			&\includegraphics[height=3.3cm]{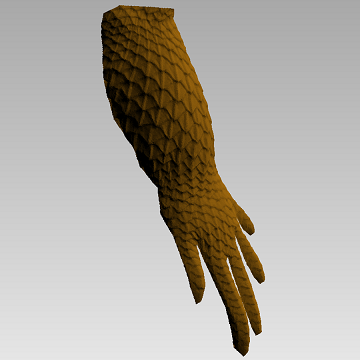}
			&\includegraphics[height=3.3cm]{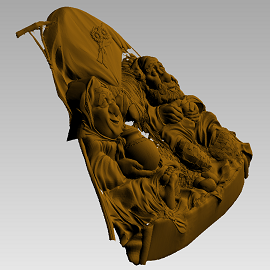}
			&\includegraphics[height=3.3cm]{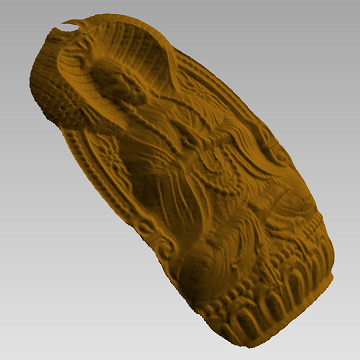}\\		
			\includegraphics[height=3.3cm]{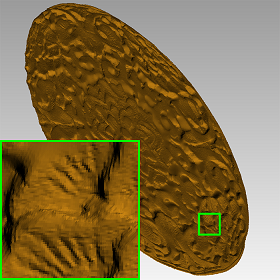}
			&\includegraphics[height=3.3cm]{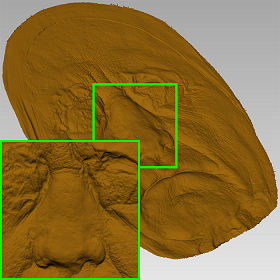}
			&\includegraphics[height=3.3cm]{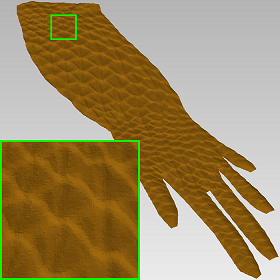}
			&\includegraphics[height=3.3cm]{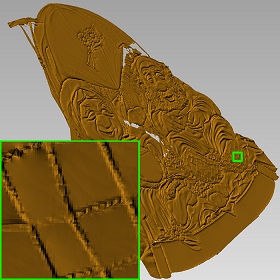}
			&\includegraphics[height=3.3cm]{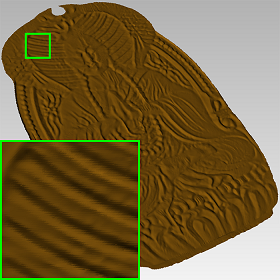}\\
\includegraphics[width=0.181 \textwidth]{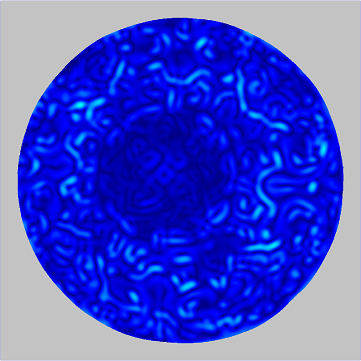}
&\includegraphics[width=0.181 \textwidth]{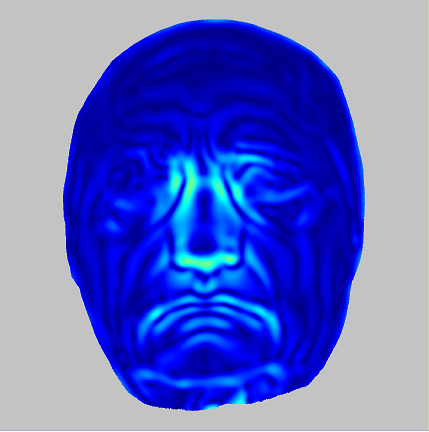}
&\includegraphics[width=0.181 \textwidth]{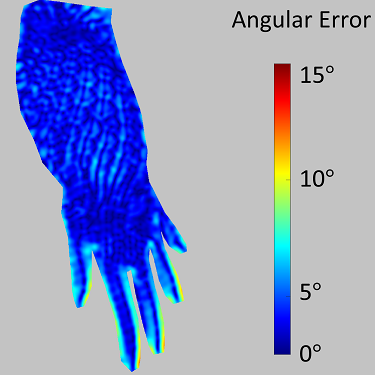}
&\includegraphics[width=0.181 \textwidth]{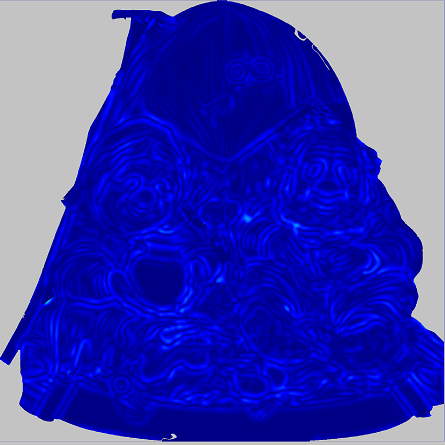}
&\includegraphics[width=0.181 \textwidth]{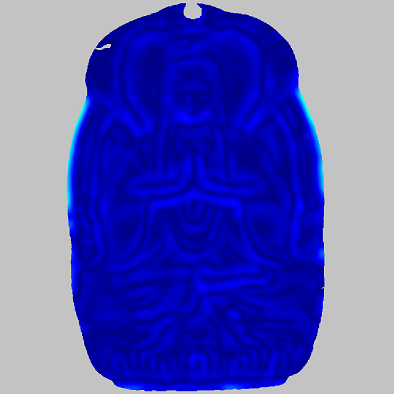}\\
\footnotesize{\textit{Circular} ($0.96^\circ$)}  &\footnotesize{\textit{Geothe} ($0.95^\circ$)}  &\footnotesize{\textit{Lizard} ($1.47^\circ$)}  &\footnotesize{\textit{Panno} ($0.44^\circ$)}  &\footnotesize{\textit{Woodcarving} ($0.47^\circ$)}\\
\end{tabular} 
		\caption{\textbf{Property of the extracted \textit{detail component} representation.} The top row shows five five different 3D models, the middle row shows the reconstructed surface of the corresponding \textit{shape component}, and the bottom row shows the error map $E_{mae}$ between $\Bar{\Delta}_{\mathcal{N}}$ and $\mathbf{z}$. The reconstructed surfaces of the \textit{detail component} maps are flatten and oriented towards the $z$-axis, which is shape-uncorrelative to the original surface.}
\label{fig_property_detail} 
\end{figure*}

For a surface with the normal map $\mathcal{N}$, its surface shape in the normal domain is represented as $\bar {\mathcal{N}}$ and formulated by Eq. \eqref{eq_conv_cal}.
\begin{equation}\label{eq_conv_cal}
\small
\bar {\mathcal{N}}\left( {r,c} \right) = Conv \left( \mathcal{N}, \kappa   \right),
\end{equation}
where $\kappa$ denotes a $(2w+1)$ by $(2w+1)$  \textit{Gaussian} or average convolution kernel, $w$ is set to $5$ or $7$ in the experiments, and $Conv (\cdot)$ represents the convolution operation.

It is noted that all the discussed normal vectors are {normalized below, and} can be represented  on a unit sphere. $\bar {\mathcal{N}}\left( {r,c} \right) \in \mathbb{R} ^3$ in Eq. (\ref{eq_conv_cal}) can be visualized as the center of a solid angle $\eta(r,c)$, which is a subset of normals in a {$(2w+1)$$\times$$(2w+1)$ window}, \textit{i.e.},  $\eta(r,c)=\left\{ {\mathcal{N}\left( {s,t} \right)\left| {r - w \le s \le r + w;c - w \le t \le t + w} \right.} \right\}$. Apparently, the smoother the normal map $\mathcal{N}$, the smaller the variance of the normals in $\eta(\cdot)$, and the closer $\mathcal{N}(r,c)$ to the center of $\eta(\cdot)$. In turn, it can be considered that $\bar{\mathcal{N}}$ represents the pure shape of $\mathcal{N}$. Therefore, $\bar {\mathcal{N}}$ in Eq. (\ref{eq_conv_cal}) is also called \textit{shape component}, and $\bar{\mathcal{N}}\left( {r,c} \right)$ is supposed to be similar to each other within a small  $(2w+1)$$\times$$(2w+1)$ window. 

\subsection{Geometry Detail Representation}\label{sub:detailDef}
Intuitively, the difference between a normal map and its \textit{shape component} determines the roughness of an input surface. This phenomenon has been verified in \cite{tasdizen2002geometric,chen2005sharpness,higo2009hand}, where a normal map can be sharpened by increasing the angle between $\mathcal{N}$ and $\bar{\mathcal{N}}$. In this section, we use this clue to define \textit{detail component} in terms of normals.

Ideally, the \textit{detail component} feature is expected to contain no shape information. In other words, it should be only related texture or micro structure, and should not exhibit concrete shapes compared to the \textit{shape component} extracted from a surface.  Meanwhile, the \textit{detail component} should keep high similarity with the local original surface, as the detail feature can be well represented. Although the direct subtraction of $\mathcal{N}$ and $\bar{\mathcal{N}}$ can describe this subtle difference, it will be over strict when used to measure the practical geometry details. Therefore, how to derive a {detail-without-shape} feature is our focus.

Let us consider a normal vector $\mathcal{N}(r,c)$ and its \textit{shape component} $\bar {\mathcal{N}}\left( {r,c} \right)$ as given in Fig. \ref{fig_relativeCoord} (left), in which our interests are primarily focused on their relative differences. A reasonable way to describe such a relative difference is to transform them into a relative coordinate system, under which, the mapping of $\mathcal{N}(r,c)$ can represent this difference as illustrated in Fig. \ref{fig_relativeCoord} (right). By comparing these two coordinate systems in Fig. \ref{fig_relativeCoord}, it is not difficult to find that such a relative coordinate system takes the mapping of $\bar {\mathcal{N}}\left( {r,c} \right)$ as its $z$-axis. To this end, we define the \textit{detail component} along with the relative coordinate transformation as follows.

For a given $\mathcal{N}(r,c)$ and its \textit{shape component} $\bar{\mathcal{N}}(r,c)$, the corresponding \textit{detail component} ${\Delta _\mathcal{N}}\left( {r,c} \right)$ can be computed as the mapping of $\mathcal{N}(r,c)$ in the relative coordinate system whose $z$-axis is the mapping of $\bar{\mathcal{N}}(r,c)$.
\begin{equation}\label{eq_detail_form}
	{\Delta _\mathcal{N}}\left( {r,c} \right) = \mathbf{R}\left| {_{\bar {\mathcal{N}}\left( {r,c} \right) , \mathbf{z}}} \right. \cdot \mathcal{N}\left( {r,c} \right),
\end{equation}
where $\cdot$ denotes the multiplication operator of matrix and vector, $\mathcal{N}(r,c)$ is a 3$\times$1 column vector, and $\mathbf{R}\left| {_{\bar {\mathcal{N}}\left( {r,c} \right) , \mathbf{z}}} \right.$ represents a 3$\times$3 rotation matrix that can transform vector $\bar {\mathcal{N}}\left( {r,c} \right)$ to the vector  $\mathbf{z}=[0,0,1]^T$, \textit{i.e.},  
\begin{equation}
\label{eq_z_and_shape}
\mathbf{z}=\mathbf{R}\left| {_{\bar {\mathcal{N}}\left( {r,c} \right) , \mathbf{z}}} \right. \cdot  {\bar {\mathcal{N}}\left( {r,c} \right)}.
\end{equation}
  According to Rodrigues' rotation formula \cite{gallego2015compact}, $\mathbf{R}\left| {_{\bar {\mathcal{N}}\left( {r,c} \right), \mathbf{z}}} \right.$ can be determined from $\bar{\mathcal{N}}(r,c)$ and $\mathbf{z}$ as	
\begin{equation}\label{eq_detail_rotForm}
\small
	\mathbf{R}\left| {_{{\bar {\mathcal{N}}}\left( {r,c} \right),\mathbf{z}}} \right. = \mathbf{I} + {\left\lfloor {{\bar {\mathcal{N}}}\left( {r,c} \right) \otimes \mathbf{z}} \right\rfloor _ \times } + \frac{{\left\lfloor {{\bar {\mathcal{N}}}\left( {r,c} \right) \otimes \mathbf{z}} \right\rfloor _ \times ^2}}{{1 + {\bar {\mathcal{N}}}\left( {r,c} \right) \odot \mathbf{z}}},
\end{equation}
where {$\otimes$ denotes the cross product operator, $\odot$ represents the dot product operator}, $\mathbf{I} \in \mathbb{R}^{3 \times 3}$ denotes the identity matrix, and $\left \lfloor {\mathbf{v}} \right \rfloor _{\times}$ represents the skew-symmetric cross-product matrix of vector $\mathbf{v}$.
Finally, we obtain the expression of $\mathbf{R}\left| {_{\bar {\mathcal{N}}\left( {r,c} \right) , \mathbf{z}}} \right.$ as,
\begin{equation}\label{eq_detail_rotCal}
\mathbf{R}\left| {_{\bar {\mathcal{N}}\left( {r,c} \right) , \mathbf{z}}} \right. = \begin{bmatrix}
1-\frac{\bar{n}_x^2}{1+\bar{n}_z} & - \frac {\bar{n}_x\bar{n}_y}{1+\bar{n}_z} & -\bar{n}_x \\ 
- \frac {\bar{n}_x\bar{n}_y}{1+\bar{n}_z} & 1-\frac{\bar{n}_y^2}{1+\bar{n}_z} & -\bar{n}_y  \\ 
\bar{n}_x & \bar{n}_y & 1-\frac{\bar{n}_x^2+\bar{n}_y^2}{1+\bar{n}_z}
\end{bmatrix},
\end{equation}
where $\bar{n}_x$, $\bar{n}_y$ and $\bar{n}_z$ are the $xyz$ components of ${\bar {\mathcal{N}}}\left( {r,c} \right)$.

Fig. \ref{fig_property_detail} shows five representative 3D models and the relevant \textit{detail component} results. It is noted that, to visualize the \textit{detail component}, we obtain their depth values via the \textit{surface from normals} (SfN) method in \cite{xie2014surface}. In addition, we also measure the shape of $\Delta_{\mathcal{N}}$, where the energy map of angular error between $\Bar{\Delta}_{\mathcal{N}}$ and $\mathbf{z}$ is calculated. Moreover, we calculate the structure similarity (SSIM) metric    \cite{wang2004image} between the extracted \textit{detail component} by Eq. (\ref{eq_detail_form}) and its original normal map.  From left to right, the relevant $\mathbf{E}_{ssim}$ scores are 0.9566, 0.9531, 0.9882, 0.9557, and 0.9737, respectively. 
As expected, all the extracted \textit{detail component} is {detail-without-shape} and appears to be unwrapped and flattened.

\section{Properties of the Detail Component} 
\label{sec:detailproperty}
The \textit{detail component} is not just an unwrapped surface as it looks. To better analyze the characteristics of the \textit{detail component},  we demonstrate three important properties and provide the related mathematical derivations as follows.

\subsection{Detail Separability}
\label{subsubsec:separability}
According to the \textit{detail component} extracted by Eq. (\ref{eq_detail_form}), we find that the shape of an \textit{detail component} is a flatten surface and oriented towards the $z$-axis. In other words, the \textit{detail component} contains no shape information compared to the source 3D surface.  This property of the \textit{detail component} is named as \textit{detail separability}.  We give a detailed explanation as follows. 
\begin{figure*}[!t]
\centering
\begin{tabular}{p{2.4cm}<{\centering} p{2.4cm}<{\centering} p{2.4cm}<{\centering} p{2.4cm}<{\centering} p{2.4cm}<{\centering} p{2.4cm}<{\centering}}
			\includegraphics[height=2.7cm]{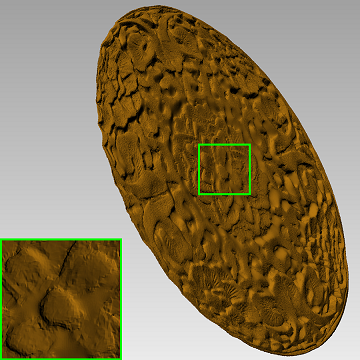}
			&\includegraphics[height=2.7cm]{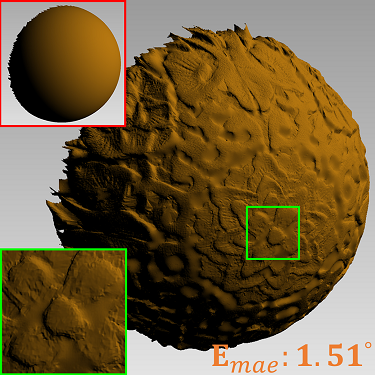}
			&\includegraphics[height=2.7cm]{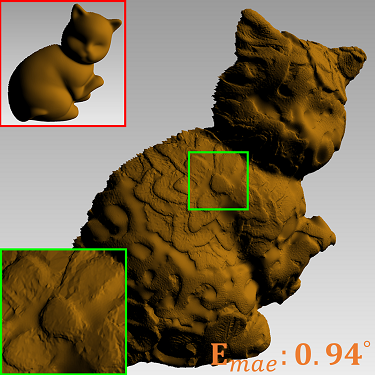}
			&\includegraphics[height=2.7cm]{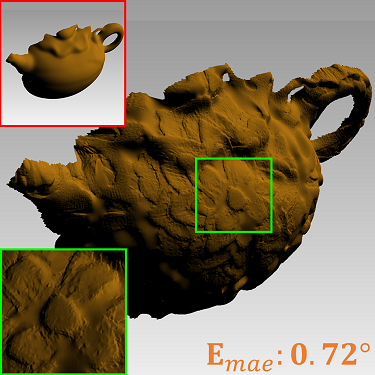}
			&\includegraphics[height=2.7cm]{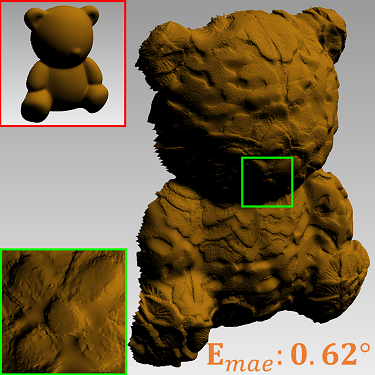}
			&\includegraphics[height=2.7cm]{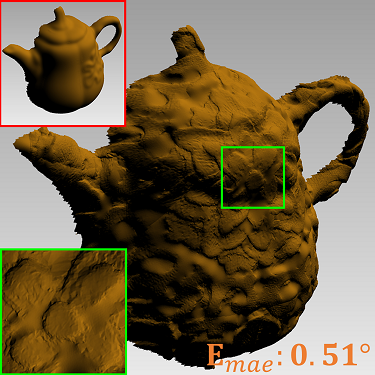}\\
			\small{\textit{Original details}} &\small{\textit{Ball}} & \small{\textit{Cat}} & \small{\textit{Pot1}} & \small{\textit{Bear}} & \small{\textit{Pot2}} \\
			\includegraphics[height=2.7cm]{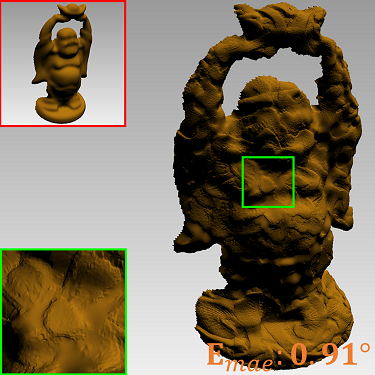}
			&\includegraphics[height=2.7cm]{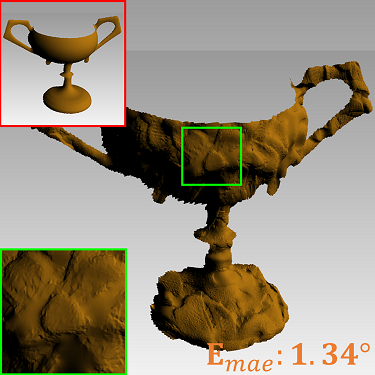}
			&\includegraphics[height=2.7cm]{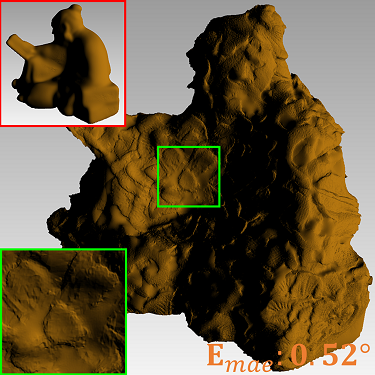}
			&\includegraphics[height=2.7cm]{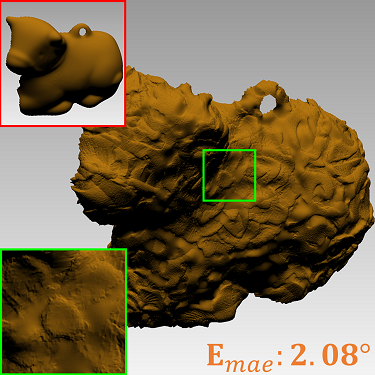}
			&\includegraphics[height=2.7cm]{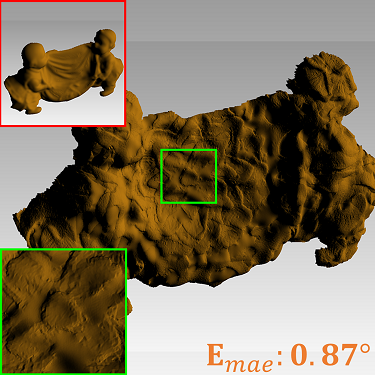}
			&\includegraphics[height=2.7cm]{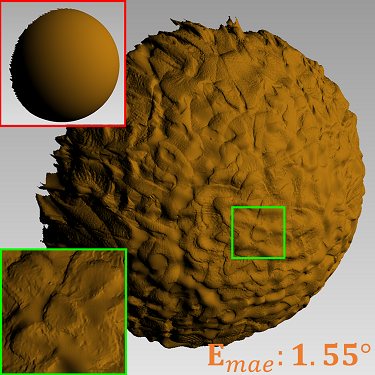}\\
			\small{\textit{Buddha}} &\small{\textit{Goblet}} & \small{\textit{Reading}} & \small{\textit{Cow}} & \small{\textit{Harvest}} & \small{$^{*}$\textit{Ball}} \\
\end{tabular} 
\caption{ \textbf{Illustration of the \textit{detail transferability} property for the proposed  \textit{detail component}.} The geometry details of \textit{Circular} is transferred one after another on 10 different models from DiLiGenT \cite{shi2019benchmark} and finally transferred back to the first model ($^{*}$\textit{Ball}). The shape difference (in terms of mean angular error (MAE)) is provided at the bottom of each   target object, which is calculated between the current shape before detail transfer and the extracted shape after detail transfer.}
\label{fig_property_transfer} 
\end{figure*}

According to the \textit{shape component} definition in Eq. (\ref{eq_conv_cal}), $\bar{\mathcal{N}} (r,c)$ can be represented by  
\begin{equation}\label{eq_prove_uniformSum_1}
\small
\bar {\mathcal{N}} \left( {r,c} \right) = {\lambda _1}\sum\limits_{s =  - w}^w {\sum\limits_{t =  - w}^w {\kappa \left( {s,t} \right)\mathcal{N}\left( {r + s,c + t} \right)} },
\end{equation}
{ where according to Eq. \eqref{eq_z_and_shape}, $\bar{\mathcal{N}} (r,c)$ can be calculated from $\mathbf{z}$, like $\bar{\mathcal{N}}\left( {r,c} \right) = {\left( {\mathbf{R}\left| {_{\Bar{\mathcal{N}} \left( {r,c} \right),{\mathbf{z}}}} \right.} \right)^{ - 1}} {\cdot {\mathbf{z}}}$, and according to Eq. \eqref{eq_detail_form}, ${\mathcal{N}}(r,c)$ can be calculated from $\Delta_{\mathcal{N}}(r,c)$, like $\mathcal{N}\left( {r + s,c + t} \right) = {\left( {\mathbf{R}\left| {_{\Bar{\mathcal{N}} \left( {r + s,c + t} \right),{\mathbf{z}}}} \right.} \right)^{ - 1}} \cdot {\Delta _{\mathcal{N}}}\left( {r + s,c + t} \right)$. It is noted that to simplify the expression, $\lambda_1$  represents a normalization constant to ensure that the filtered normals have unit length.}

By submitting them into Eq. (\ref{eq_prove_uniformSum_1}), we have
\begin{equation}\label{eq_prove_uniformSum_2}
\small
\begin{split}
        &\left( {{{\left( {\mathbf{R}\left| {_{\Bar{\mathcal{N}} \left( {r,c} \right),{\mathbf{z}}}} \right.} \right)}^{ - 1}} \cdot {\mathbf{z}}} \right) ={\lambda _1}\\
        & \resizebox{0.90\hsize}{!}{$\sum\limits_{s =  - w}^w {\sum\limits_{t =  - w}^w {\kappa \left( {s,t} \right)\left[ {{{\left( {\mathbf{R}\left| {_{\Bar{\mathcal{N}} \left( {r + s,c + t} \right),{\mathbf{z}}}} \right.} \right)}^{ - 1}} \cdot {\Delta _{\mathcal{N}}}\left( {r + s,c + t} \right)} \right]} } $} ,
\end{split}
\end{equation}

For a dense surface, its \textit{shape component} is assumed to be continuous and change slowly within a small surface patch (the related proof can be found in \cite{malzbender2006surface}, 11$\times$11), and their corresponding rotation matrices $ {\mathbf{R}\left| {_{\bar {\mathcal{N}}\left( {r + s,c + t} \right),\mathbf{z}}} \right.} $ should be similar. {To verify this condition, we calculate the average rotation angle between $\left. \mathbf{R} \right|_{\bar {\mathcal{N}} \left( {r,c} \right),\mathbf{z}}$ and ${\left. \mathbf{R} \right|_{\bar {\mathcal{N}} \left( {r + s,c + t} \right),\mathbf{z}} }$ as formulated in Eq. (\ref{eq_similar_coe}).}
\begin{equation}\label{eq_similar_coe}
\small
	\bar \theta \left( {r,c} \right) = \frac{1}{{{{\left( {2w + 1} \right)}^2}}}\sum\limits_{s =  - w}^w {\sum\limits_{t =  - w}^w {\theta \left( {s,t} \right)} },
\end{equation}
where $\theta \left( \cdot \right)$ is calculated by
\begin{equation}
\small
\nonumber
\resizebox{0.95\hsize}{!}{$\theta \left( {r,c} \right) = arccos\left( {{{\left( {trace\left( {\left. \mathbf{R} \right|_{\bar {\mathcal{N}} \left( {r,c} \right),\mathbf{z}} \cdot \left( {\left. \mathbf{R} \right|_{\bar {\mathcal{N}} \left( {r + s,c + t} \right),\mathbf{z}} } \right)^T } \right) - 1} \right)} \mathord{\left/
 {\vphantom {{\left( {trace\left( {\left. \mathbf{R} \right|_{\bar {\mathcal{N}} \left( {r,c} \right),z} \left( {\left. \mathbf{R} \right|_{\bar {\mathcal{N}} \left( {r + s,c + t} \right),\mathbf{z}} } \right)^T } \right) - 1} \right)} 2}} \right.
 \kern-\nulldelimiterspace} 2}} \right)
 $},
\end{equation}
{where $T$ represents the transpose of a matrix. Apparently, the closer $\bar \theta$ to $0$, the more similar in $ {\mathbf{R}\left| {_{\bar {\mathcal{N}}\left( {r + s,c + t} \right),\mathbf{z}}} \right.}$.}

Moreover, we provide three different norms to measure the distance between \scalebox{1.0}{$ { {{\mathbf{R}}\left| {_{\Bar{\mathcal{N}} \left( {r,c} \right),{\mathbf{z}}}} \right.} \cdot {{\left( {{\mathbf{R}}\left| {_{\Bar{\mathcal{N}} \left( {r + s,c + t} \right),{\mathbf{z}}}} \right.} \right)}^{ - 1}}} $} and  the identity matrix ${\mathbf{I}}$:
\begin{equation}
\nonumber
\label{eq_similar_norm}
\begin{split}
    {\Bar{\mathcal{\ell}}_p}\left( {r,c} \right) = 
    &\resizebox{0.74\hsize}{!}{$\frac{1}{{{{\left( {2w + 1} \right)}^2}}}{\sum\limits_{s =  - w}^w {\sum\limits_{t =  - w}^w {{{\left\| {\left( {{\mathbf{R}}\left| {_{\Bar{\mathcal{N}} \left( {r,c} \right),{\mathbf{z}}}} \right.} \right) \cdot {{\left( {{\mathbf{R}}\left| {_{\Bar{\mathcal{N}} \left( {r + s,c + t} \right),{\mathbf{z}}}} \right.} \right)}^{ - 1}} - {\mathbf{I}}} \right\|}_p}} } }$},\\
    &p = 1,2,or,\infty,
\end{split}
\end{equation}
Table \ref{tab_similar_coe} provides the results of five representative 3D objects in terms of the average rotation angle $\bar \theta$ and the average value of three norms ${\Bar{\mathcal{\ell}}_p}$. As seen, all of these metrics have small values. For example, the average rotation angles are less than $0.61^\circ$, and the largest norm value is less than $0.014$.

\begin{table}[!t]
\centering 
\caption{Property of the \textit{shape component} in terms of the rotation angle and the matrix norm. } 
\small
\renewcommand\arraystretch{1.2}
\setlength{\tabcolsep}{0.9mm}{
\begin{tabular}{cccccc}
\hline\hline 
{} & {\makecell[c]{\textit{Circular}}} & {\makecell[c]{\textit{Goethe} }} & {\makecell[c]{\textit{Lizard} }} & {\makecell[c]{\textit{Panno}}} & {\makecell[c]{\textit{Woodcarving}}}\\				
\hline
{$\bar\theta$} & $0.4950^\circ$   & $0.6049^\circ$                     & $0.3249^\circ$                     & $0.3391^\circ$                    & $0.4678^\circ$\\	
${\Bar{\ell}_1}$ & 0.0109   & 0.0133                     & 0.0066                     & 0.0074                    & 0.0100                            \\
${\Bar{\ell}_2}$ & 0.0086   & 0.0106                     & 0.0057                     & 0.0059                    & 0.0082                          \\
${\Bar{\ell}_{\infty}}$ & 0.0109   & 0.0133                     & 0.0066                     & 0.0074                    & 0.0100  \\				
\hline
\hline  	
\end{tabular}}\label{tab_similar_coe}
\end{table}

As a result, every $\mathbf{R}\left| {_{\bar {\mathcal{N}}\left( {r + s,c + t} \right),\mathbf{z}}} \right.$ can be approximately replaced by $\mathbf{R}\left| {_{\bar {\mathcal{N}}\left( {r,c} \right),\mathbf{z}}} \right.$, and  Eq. (\ref{eq_prove_uniformSum_2}) can be rewritten as
\begin{equation}
\small
\begin{split}
        &\left( {{{\left( {\mathbf{R}\left| {_{\Bar{\mathcal{N}}\left( {r,c} \right),{\mathbf{z}}}} \right.} \right)}^{ - 1}} \cdot {\mathbf{z}}} \right) \\
        &\resizebox{0.90\hsize}{!}{${\approx {\lambda _1}\sum\limits_{s =  - w}^w {\sum\limits_{t =  - w}^w {\kappa \left( {s,t} \right)\left[ {{{\left( {\mathbf{R}\left| {_{\Bar{\mathcal{N}} \left( {r,c} \right),{\mathbf{z}}}} \right.} \right)}^{ - 1}} \cdot {\Delta _{\mathcal{N}}}\left( {r + s,c + t} \right)} \right]} } }$}\\
        &{ = {\lambda _1}{{\left( {\mathbf{R}\left| {_{\Bar{\mathcal{N}} \left( {r,c} \right),{\mathbf{z}}}} \right.} \right)}^{ - 1}} \cdot \sum\limits_{s =  - w}^w {\sum\limits_{t =  - w}^w {\kappa \left( {s,t} \right){\Delta _{\mathcal{N}}}\left( {r + s,c + t} \right)} } },
\end{split}
\label{eq:shape_middle}
\end{equation}
where the right side is actually the \textit{shape component} representation of ${\Delta _{\mathcal{N}}}\left( {r,c} \right)$ according to {Eq. (\ref{eq_prove_uniformSum_1}).  Finally, when multiplying ${\mathbf{R}\left| {_{{\bar {\mathcal{N}}}\left( {r,c} \right),\mathbf{z}}} \right.}$} on both side of Eq. \eqref{eq:shape_middle}, we have
\begin{equation}\label{eq_prove_uniformSum}
\mathbf{z} {\approx }  {{\bar \Delta }_{\mathcal{N}}}\left( {r,c} \right).
\end{equation}

Eq. (\ref{eq_prove_uniformSum}) reveals that the shape of ${\Delta _{\mathcal{N}}}$ is a flat surface, and oriented towards the $z$-axis. Moreover, it is shape-uncorrelative to the original surface. Thus, the \textit{detail separability} property has been demonstrated.  Fig. \ref{fig_property_detail} provides five representative 3D models from simple repeating patterns to complex textures. In the second row, the reconstructed surfaces {from} all five extracted \textit{detail component} maps are oriented towards the $z$-axis, which {are} consistent with the deduction in Eq. (\ref{eq_prove_uniformSum}).

\subsection{Detail Transferability}
\label{subsubsec:transferability}	
The significance of the proposed \textit{detail component} representation lies in the fact that it can be extracted as separable features used for 3D surface editing, in which the texture transferability to another surface is an essential goal. According to Eq. (\ref{eq_detail_form}), a surface normal $\mathcal{N}\left( {r,c} \right)$ can be computed from its \textit{shape component} $\bar{\mathcal{N}}(r,c)$ and the \textit{detail component} $\Delta_{\mathcal{N}}(r,c)$,
\begin{equation}\label{eq_invert_detail_rotForm}
\small
	\begin{split}
	{\mathcal{N}}\left( {r,c} \right) &= ({\mathbf{R}}\left| {_{{\bar {\mathcal{N}}}\left( {r,c} \right),\mathbf{z}}} \right.)^{-1} \cdot {\Delta _{\mathcal{N}}}\left( {r,c} \right) \\
	&={\mathbf{R}}\left| {_{{\mathbf{z},\bar {\mathcal{N}}}\left( {r,c} \right)}} \right. \cdot {\Delta _{\mathcal{N}}}\left( {r,c} \right).
	\end{split}
\end{equation}	
\begin{figure}[!t]
\centering
\includegraphics[width=7.5cm]{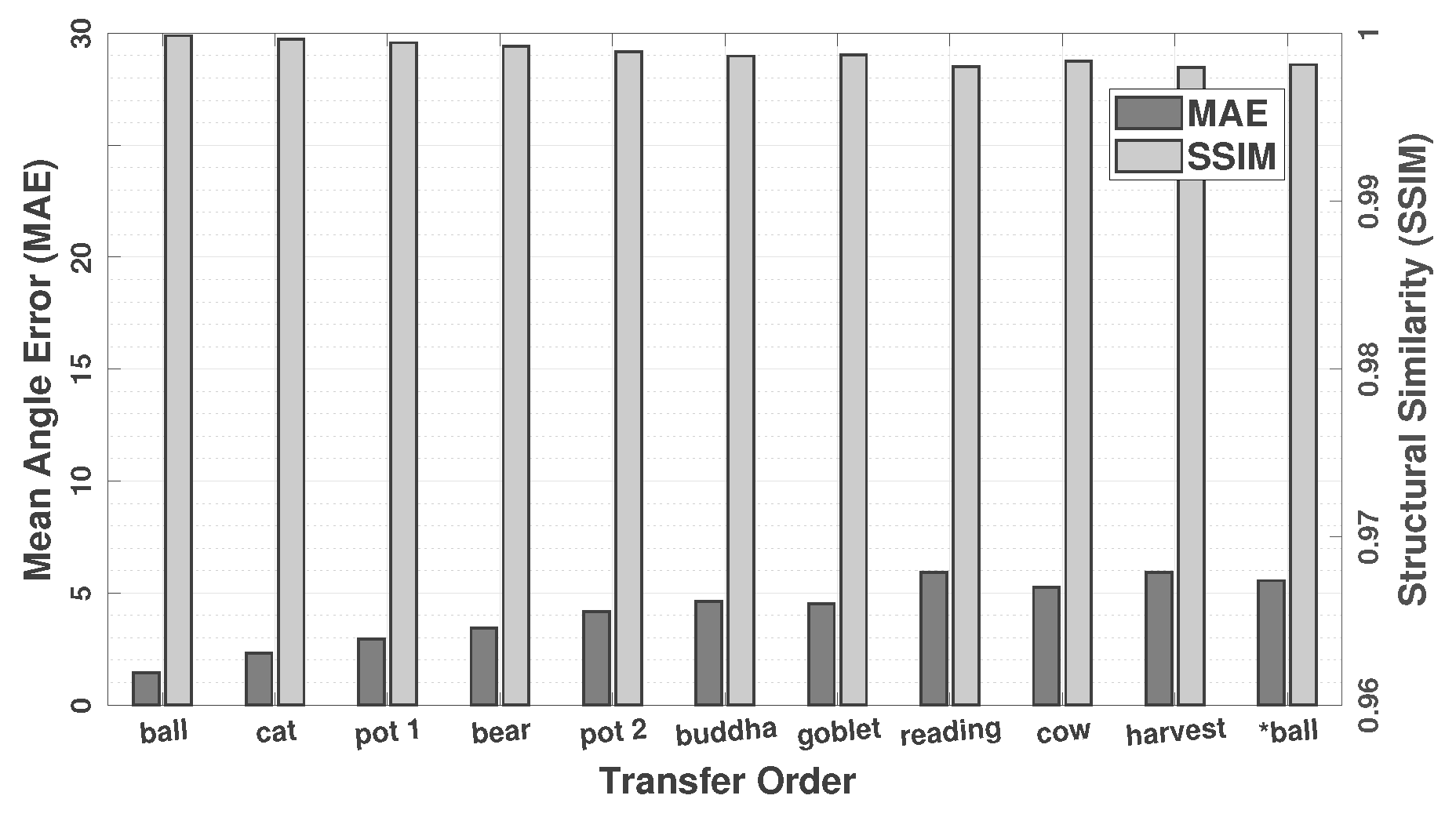}
\caption{\textbf{The \textit{mean angular error} (MAE) and \textit{structural similarity} (SSIM) results of \textit{Circular}}. The reported MAE and SSIM results are obtained by transferring one after another on $10$ different models.}
\label{fig_transBar}
\end{figure}

Considering the \textit{detail separability} property,  ${\Delta _{\mathcal{N}(r,c)}}$ contains no concrete shape information (\textit{i.e.}, a flat plane), which implies  that $\bar{\mathcal{N}}(r,c)$ can be replaced with an arbitrarily target shape. In Eq. (\ref{eq_invert_detail_rotForm}), replacing $\bar{\mathcal{N}}(r,c)$ with another surface while keeping ${\Delta _{\mathcal{N}(r,c)}}$ unchanged is named as \textit{detail transferability}. We demonstrate the \textit{detail transferability} property, and provide the related mathematical proofs as follows.

Let the \textit{detail component} ${\Delta _{{{\mathcal{N}}_{org}}}}$ be separated from {a source normal map $\mathcal{N}_{org}$}, and assume that it can be transferred to a target surface with the normal map ${\bar{\mathcal{N}}_{tgt}}$ via
\begin{equation}\label{eq_transfer}
		{{\mathcal{N}}^*}\left( {r,c} \right) = \mathbf{R}\left| {_{\mathbf{z},{{{\bar {\mathcal{N}}}}_{tgt}}\left( {r,c} \right)}} \right. \cdot {\Delta _{{{\mathcal{N}}_{org}}}}\left( {r,c} \right).
\end{equation}
For a successful \textit{geometry detail transfer}, $\mathcal{N}^*$ is expected to have the similar surface shape as that of ${\bar{\mathcal{N}}_{tgt}}$ as well as the similar geometry details as that of   $\mathcal{N}_{org}$. In the following section, we demonstrate that both of them can be guaranteed based on the proposed normal-based framework.

\noindent \textbf{Shape similarity:} According to {Eq. (\ref{eq_prove_uniformSum_1})}, the \textit{shape component} of ${{\mathcal{N}}^*}\left( {r,c} \right)$ can be represented by 
\begin{equation}\label{eq_paral_trans}
\small
    {\Bar{\mathcal{N}} ^*}\left( {r,c} \right) = {\lambda _2}\sum\limits_{s =  - w}^w {\sum\limits_{t =  - w}^w {\kappa \left( {s,t} \right){\mathcal{N}^*}\left( {r + s,c + t} \right)} },
\end{equation}
where { $\lambda_2$ represents a normalization constant to ensure that the filtered normals have unit length.}

For a smooth $\bar{\mathcal{N}}_{tgt}$ (with the shape information only), $\{\mathbf{R}\mid_{\mathbf{z},\mathcal{N}_{tgt}(r+s,c+t)}\}$ is  highly similar to each other within {a small $(2w+1)$$\times$$(2w+1)$ window}, as demonstrated in Section \ref{subsubsec:separability}. Moreover, by submitting $\mathcal{N}^* (r,c)$  into Eq. (\ref{eq_paral_trans}),  it has the following approximated expression, 
\begin{equation}
\label{eq_trans_shapeCom0}
\small
		\begin{split}
		&{{\bar {\mathcal{N}}}^*}\left( {r,c} \right)\approx \\
		&\resizebox{0.88\hsize}{!}{${\lambda _2}\mathbf{R}\left| {_{{\mathbf{z}},{{\bar {\mathcal{N}}}_{tgt}}\left( {r,c} \right)}} \right. \cdot \sum\limits_{s =  - w}^w {\sum\limits_{t =  - w}^w {\kappa \left( {s,t} \right){\Delta _{{\mathcal{N}_{org}}}}\left( {r + s,c + t} \right)} }$}.
		\end{split}
\end{equation}

According to the \textit{detail separability} property, the sum of $\Delta_{\mathcal{N}_{org}}(r,c)$ in a small window is approximately parallel to $z$-axis. Thus, Eq. (\ref{eq_trans_shapeCom0}) can be simplified as
\begin{equation}\label{eq_trans_shapeCom}
\small
	\begin{split}
	\bar{\mathcal{N}}^{*}(r, c) &\approx \mathbf{R}\mid_{\mathbf{z}, {\bar {\mathcal{N}}}_{tgt}(r, c)} \cdot ~~\mathbf{z} \\
	&= {\bar{\mathcal{N}}_{tgt}}\left( {r,c} \right).
	\end{split}
\end{equation}

Consequently, this indicates that the transferred result, $\mathcal{N}^*$, has the similar shape as the target surface $ \bar{\mathcal{N}}_{tgt}$. 
Fig. \ref{fig_property_transfer} shows the \textit{detail transferability} property on the DiLiGenT dataset \cite{shi2019benchmark}. The \textit{detail component} of the \textit{Circular} model from \textit{Aim@Shape}\footnote{http://visionair.ge.imati.cnr.it} is used as the source ${\Delta _{{{\mathcal{N}}_{org}}}}\left( {r,c} \right)$, and the \textit{shape  components} of all 3D models from  the DiLiGenT are used as the target ${\bar{\mathcal{N}}_{tgt}}\left( {r,c} \right)$. 
 The widely-used mean angular error (MAE) \cite{xie2014surface,santo2017deep} is also adopted to measure the reconstructed shape similarity.  In Fig. \ref{fig_property_transfer}, the numeric values are calculated between the target shape before detail transfer and the extracted shape after detail transfer. For instance, the $E_{mae}$ value of \textit{Bear} represents the shape difference between before and after detail transfer from \textit{Pot1}.
The maximum $\mathbf{E}_{mae}$ value is $2.08^{\circ}$, which indicates a very promising shape  similarity. The visual results are consistent with the MAE results.
\begin{figure}[!t]
\centering
\begin{tabular}{p{1.4cm}<{\centering} p{1.4cm}<{\centering} p{1.4cm}<{\centering} p{1.4cm}<{\centering} p{1.4cm}<{\centering}}
	    \includegraphics[width=1.6cm, height=1.8cm]{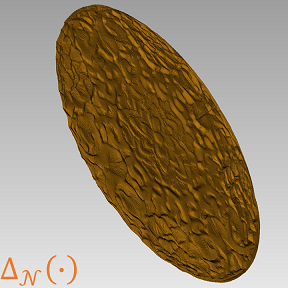}
		&\includegraphics[width=1.6cm, height=1.8cm]{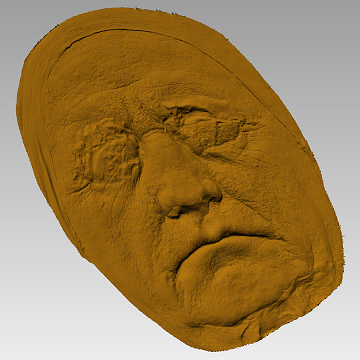}
		&\includegraphics[width=1.6cm, height=1.8cm]{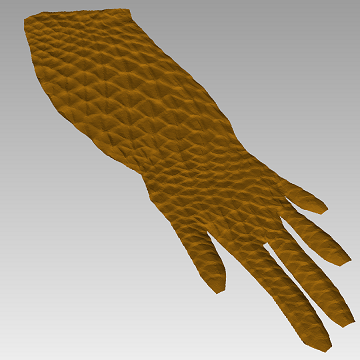}
		&\includegraphics[width=1.6cm, height=1.8cm]{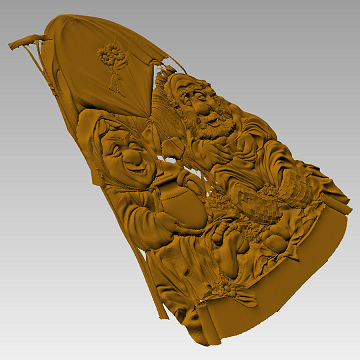}
		&\includegraphics[width=1.6cm, height=1.8cm]{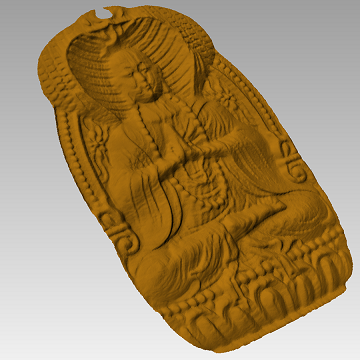}\\
		\includegraphics[width=1.6cm, height=1.8cm]{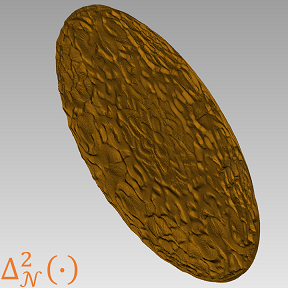}
		&\includegraphics[width=1.6cm, height=1.8cm]{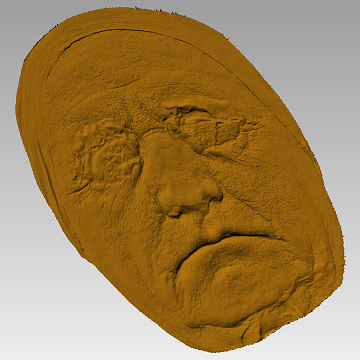}
		&\includegraphics[width=1.6cm, height=1.8cm]{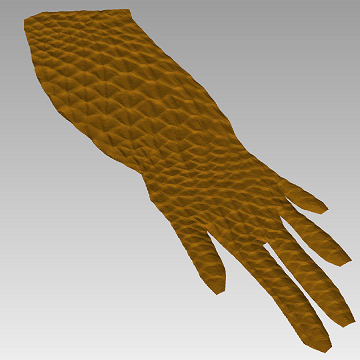}
		&\includegraphics[width=1.6cm, height=1.8cm]{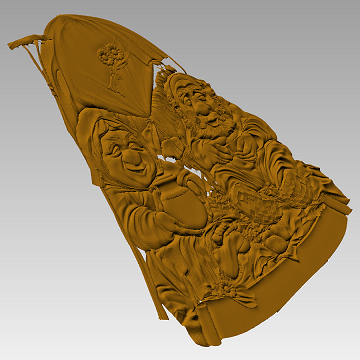}
		&\includegraphics[width=1.6cm, height=1.8cm]{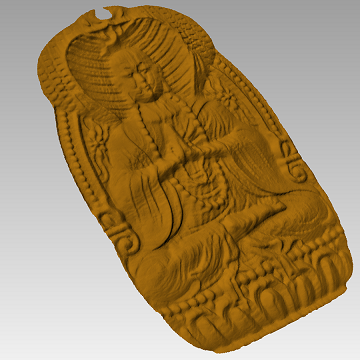}\\
		\includegraphics[width=1.6cm, height=1.8cm]{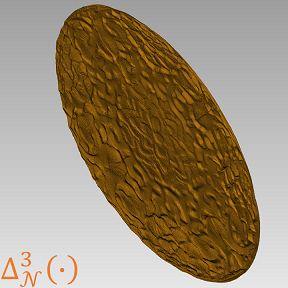}
		&\includegraphics[width=1.6cm, height=1.8cm]{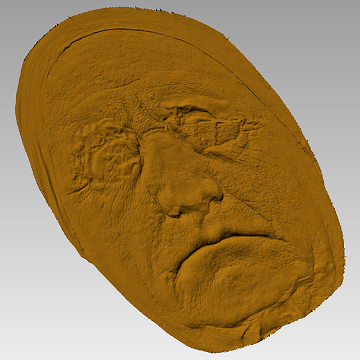}
		&\includegraphics[width=1.6cm, height=1.8cm]{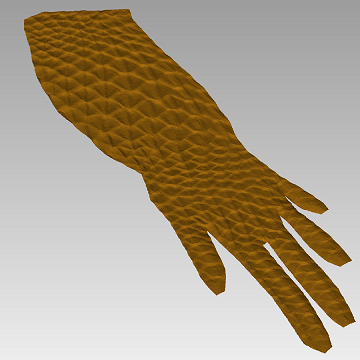}
		&\includegraphics[width=1.6cm, height=1.8cm]{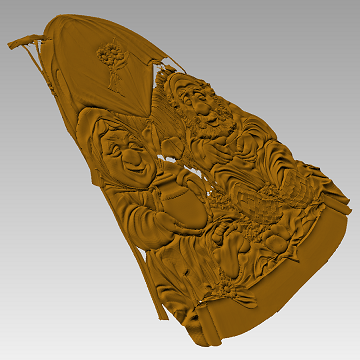}
		&\includegraphics[width=1.6cm, height=1.8cm]{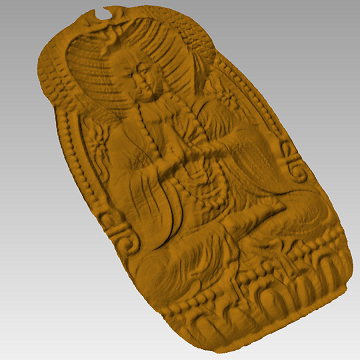}\\
		\includegraphics[width=1.6cm, height=1.8cm]{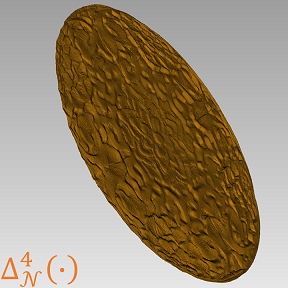}
		&\includegraphics[width=1.6cm, height=1.8cm]{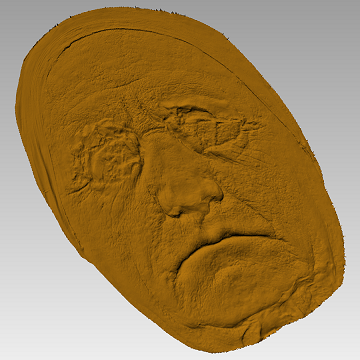}
		&\includegraphics[width=1.6cm, height=1.8cm]{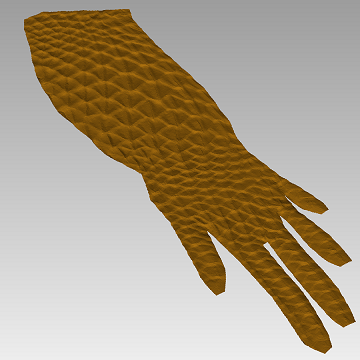}
		&\includegraphics[width=1.6cm, height=1.8cm]{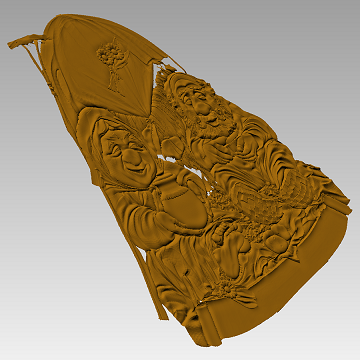}
		&\includegraphics[width=1.6cm, height=1.8cm]{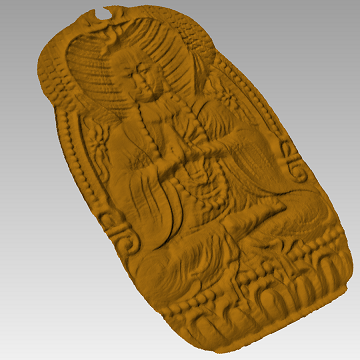}\\
		\small{{\footnotesize\textit{Circular}}} & \small{{\footnotesize\textit{Geothe}}} & 
		\small{{\footnotesize\textit{Lizard}}} & 
		\small{{\footnotesize\textit{Panno}}} &
		\small{{\footnotesize\textit{Woodcarving}}}\\
\end{tabular} 
\caption{\textbf{Illustration of the \textit{detail  idempotence} property for the \textit{detail component}.} From the top row to the bottom row: ${\Delta _{\mathcal{N}}}\left(\cdot \right)$, ${\Delta _{\mathcal{N}}^2}\left( \cdot \right)$, ${\Delta _{\mathcal{N}}^3}\left(\cdot \right)$, and ${\Delta _{\mathcal{N}}^4}\left( \cdot \right)$.
}
\label{fig_property_invariance} 
\end{figure}

\noindent \textbf{Detail similarity:} The \textit{detail component} of $\mathcal{N}^*$ can be calculated by Eq. (\ref{eq_detail_form}) as 
\begin{equation}\label{eq_trans_detailCom0}
\small
		{\Delta _{{{\mathcal{N}}^*}}}\left( {r,c} \right) = \mathbf{R}\left| {_{{{{\bar {\mathcal{N}}}}^*}\left( {r,c} \right),\mathbf{z}}} \right. \cdot {{\mathcal{N}}^*}\left( {r,c} \right).
\end{equation}

By submitting Eq. (\ref{eq_trans_shapeCom}) and Eq. (\ref{eq_transfer}) into it, we have 
\begin{equation}\label{eq_trans_detailCom}
\small
		\begin{split}
		{\Delta _{{{\mathcal{N}}^*}}}\left( {r,c} \right) &\approx \mathbf{R}\left| {_{{{{\bar {\mathcal{N}}}}_{tgt}}\left( {r,c} \right),\mathbf{z}}} \right. \cdot \mathbf{R}\left| {_{\mathbf{z},{{{\bar {\mathcal{N}}}}_{tgt}}\left( {r,c} \right)}} \right. \cdot {\Delta _{{{\mathcal{N}}_{org}}}}\left( {r,c} \right)\\
		&= {\Delta _{{{\mathcal{N}}_{org}}}}\left( {r,c} \right).
		\end{split}
\end{equation}

This indicates that {the transferred result of $\mathcal{N}^*$ has the similar detail as the original surface of $\mathcal{N}_{org}$}. Fig. \ref{fig_property_transfer} shows the experiments of the \textit{detail component} transferred on various 3D surface shapes. To evaluate the robustness of {detail similarity} in {Eq. (\ref{eq_trans_detailCom})},  we evaluate the attenuation of the \textit{detail component} of \textit{Circular} which is transferred and re-extracted one after another on $10$ different models from the \textit{DiLiGenT}. For example, the \textit{detail component} of the \textit{Circular} is transferred on the first model, \textit{i.e.}, \textit{Ball}. Then, the new \textit{detail component} is re-extracted from the transferred result of \textit{ball}, and it is transferred on the second model, \textit{i.e.}, \textit{Cat}. Repetitively, this operation is performed on 10 different 3D models, and finally, it is re-transferred on the first model denoted as $^{*}$\textit{Ball}.
\begin{figure}[!t]
\centering
\includegraphics[width=0.97\linewidth]{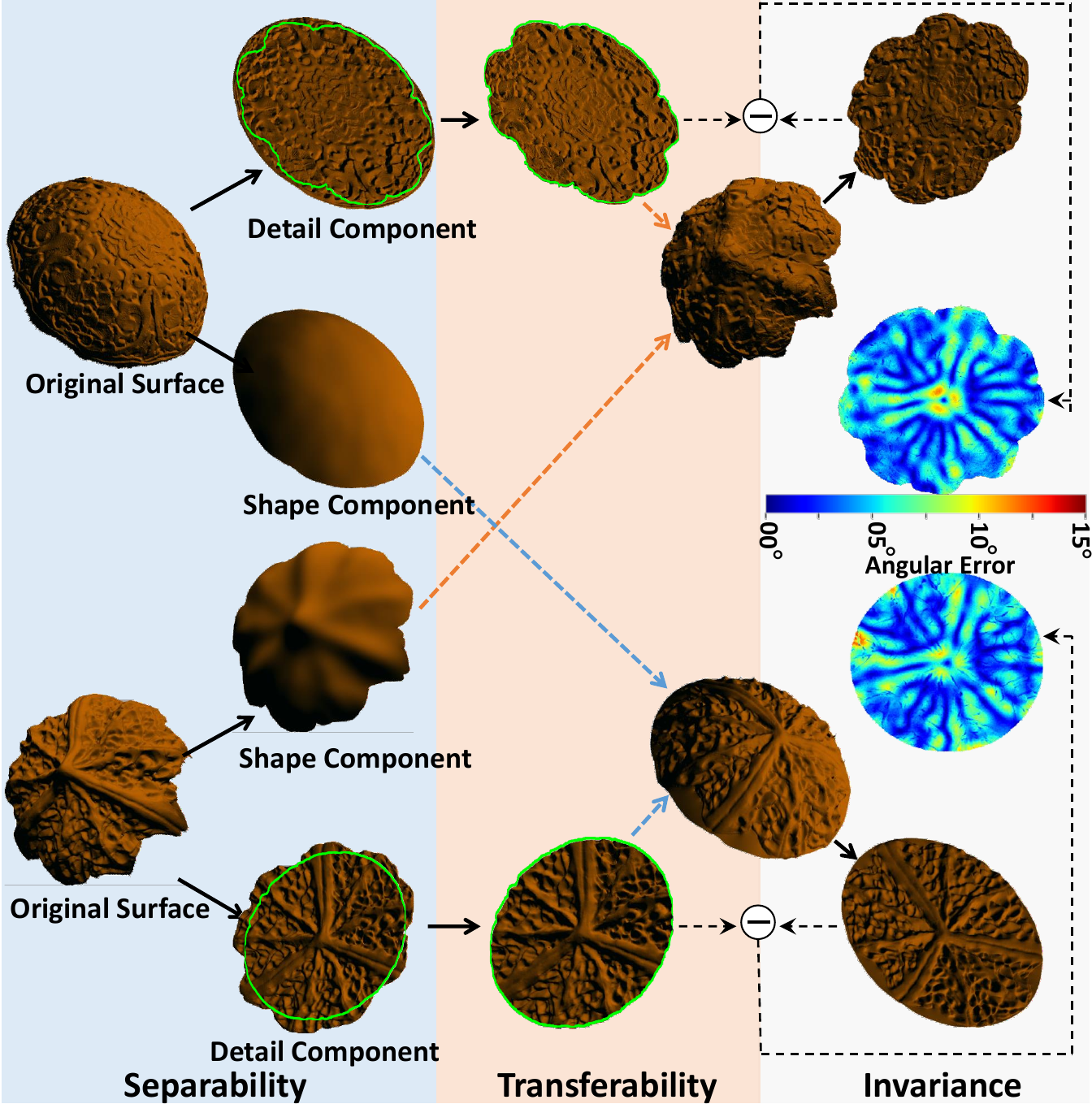}
\caption{\textbf{Illustration of the three properties for the \textit{detail component}.} The \textit{Circular} and \textit{Walnut} models are used to demonstrate the three properties of the \textit{detail component}, including separability, transferability, and  idempotence as described in Sections \ref{subsubsec:separability},  \ref{subsubsec:transferability}, and  \ref{subsubsec:invariance}, respectively. }
\label{fig_threeProperties}
\end{figure}

In addition, we evaluate the detail similarity of Fig. \ref{fig_property_transfer} in terms of both $\mathbf{E}_{ssim}$ and $\mathbf{E}_{mae}$ 
{ from the perspective of accumulation transfer error.  In Fig. \ref{fig_transBar}, the numeric values are calculated between the transferred detail exacted from the target object and that from the original  one.  For instance, the $E_{mae}$ value of \textit{Bear} represents the exacted detail difference between \textit{Bear} and \textit{Ball}.} 
The bar-chart in Fig. \ref{fig_transBar} shows that,  with the increase of the transferred times, the detail similarity decays slowly, which is consistent with our cognition. At the same time, the $\mathbf{E}_{ssim}$ values are above $0.9980$, and the $\mathbf{E}_{mae}$ values are below $5^\circ$.

\subsection{Detail  Idempotence}
\label{subsubsec:invariance}
In this section, {we demonstrate that} the repeatedly extracted \textit{detail component} is highly similar to the original \textit{detail component} map, also called \textit{detail  idempotence}.

Taking ${\Delta _{\mathcal{N}}}$ as an independent normal map, its \textit{detail component}, ${\Delta _{{\Delta _{\mathcal{N}}}}}$,  can be further expanded as:
\begin{equation}\label{eq_del_simi0}
		 {\Delta _{\mathcal{N}}^2}\left( {r,c} \right) \buildrel \Delta \over=   {\Delta _{{\Delta _{\mathcal{N}}}}}\left( {r,c} \right) = \mathbf{R}\left| {_{{{\bar \Delta }_{\mathcal{N}}}\left( {r,c} \right),\mathbf{z}}} \right. \cdot {\Delta _{\mathcal{N}}}\left( {r,c} \right),
\end{equation}
where ${{{\bar \Delta }_{\mathcal{N}}}\left( {r,c} \right)}$ can be approximated by $z$, as demonstrated in Section \ref{subsubsec:separability}. Fig. \ref{fig_property_invariance} provides the \textit{detail component} results of the first four orders ${\Delta _{\mathcal{N}}^k}\left( {r,c} \right),  k=1,2,3,4$.

Since $\mathbf{z} = \lambda_1 \bar \Delta _{{\mathcal{N}}} \left( {r,c} \right)$, $\mathbf{R}\left| {_{{{\bar \Delta }_{\mathcal{N}}}\left( {r,c} \right),\mathbf{z}}} \right.$ can be approximated as $\mathbf{R}\left| {_{\mathbf{z},\mathbf{z}} } \right.$, which is an identity matrix according to Rodrigues' rotation formula \cite{gallego2015compact}. {Therefore, $\mathbf{R}\left| {_{{{\bar \Delta }_{\mathcal{N}}}\left( {r,c} \right),\mathbf{z}}} \right.$ is close to an identity matrix, and Eq. (\ref{eq_del_simi0}) can have a similar form as follows.
\begin{equation}\label{eq_del_simi}
		{\Delta _{\mathcal{N}}^2}\left( {r,c} \right) \approx {\Delta _{\mathcal{N}}}\left( {r,c} \right).
\end{equation}

\begin{figure}[!t]
	\centering
	\includegraphics[width=\linewidth]{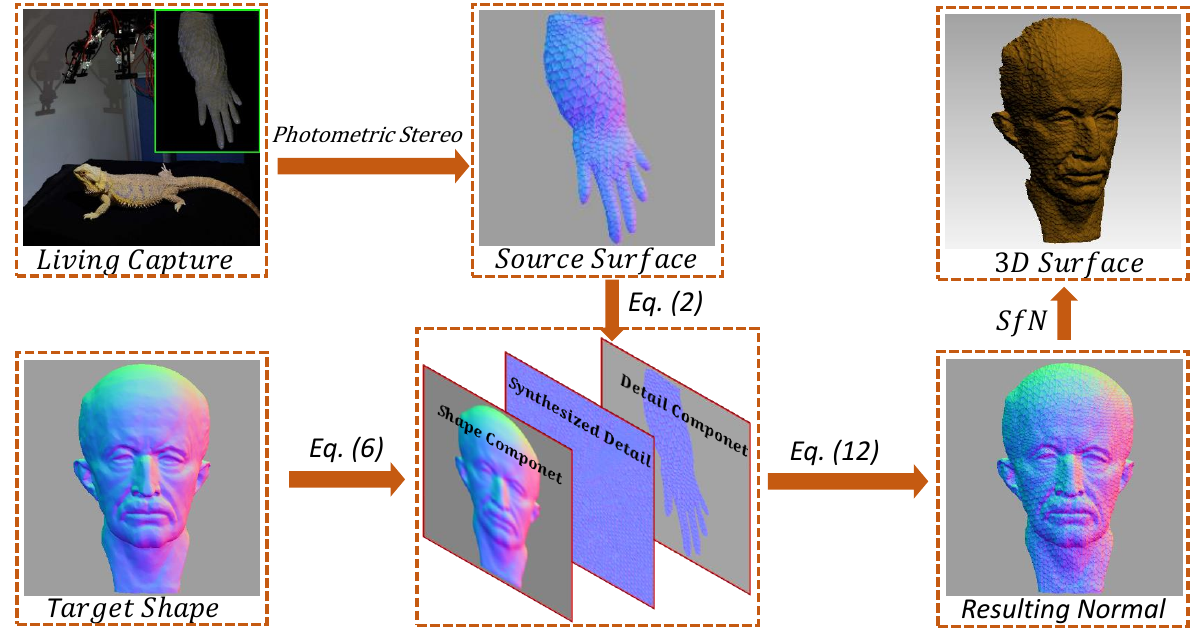}
	\caption{\textbf{Flowchart of the \textit{geometric texture synthesis}.} The \textit{shape component} and \textit{detail component} of the original normal are extracted by Eq. (\ref{eq_conv_cal}) and Eq. (\ref{eq_detail_form}), respectively. Then, the \textit{detail component} is synthesized by \cite{efros1999texture} and transferred to the target normal by Eq. (\ref{eq_transfer}). Finally, the target 3D surface is reconstructed by the surface-from-normal (SfN) method \cite{xie2014surface}.}
	\label{fig_synthesis} 
\end{figure}
\begin{table}[!t]
\centering 
\caption{Evaluations of the \textit{detail  idempotence} property between the high order $\Delta_{\mathcal{N}}$ and the corresponding original normal map.} 
\renewcommand\arraystretch{1.3}
\small
\setlength{\tabcolsep}{0.6mm}{
\begin{tabular}{cccccc}
\hline\hline 
{\footnotesize{SSIM}} & {\makecell[c]{\textit{Circular}}} & {\makecell[c]{\textit{Goethe} }} & {\makecell[c]{\textit{Lizard} }} & {\makecell[c]{\textit{Panno}}} & {\makecell[c]{\textit{Woodcarving}}}\\	
\hline
						 
$\left[{ {\Delta _{\mathcal{N}}^2}\left( \cdot \right), {\Delta _{\mathcal{N}}}\left(\cdot \right) }\right]$ & 0.9996	&0.9988	&0.9996	&0.9970	&0.9982\\
$\left[{ {\Delta _{\mathcal{N}}^3}\left( \cdot \right), {\Delta _{\mathcal{N}}}\left(\cdot \right) }\right]$ &0.9990	&0.9977	&0.9992	&0.9936	&0.9961\\
$\left[{ {\Delta _{\mathcal{N}}^4}\left( \cdot \right), {\Delta _{\mathcal{N}}}\left(\cdot \right) }\right]$ &0.9984	&0.9969	&0.9988	&0.9909	&0.9946\\				
\hline
\hline  	
\end{tabular}}\label{tab_property_detail}
\end{table}

Eq. (\ref{eq_del_simi}) indicates that the re-extracted \textit{detail component} ${\Delta _{\mathcal{N}}^2}\left( {r,c} \right)$ from ${\Delta _{\mathcal{N}}}\left( {r,c} \right)$ is similar to each other. In the same way, we can get the similar result between the \textit{k}-th order  \textit{detail component} ${\Delta _{\mathcal{N}}^k}\left( {r,c} \right)$ and ${\Delta _{\mathcal{N}}}\left( {r,c} \right)$.
Consequently, the above demonstration shows the \textit{detail  idempotence} property.

Table \ref{tab_property_detail} quantitatively evaluates  the \textit{detail  idempotence} property on five representative 3D models. The numerical results are obtained by measuring the structure similarity between the \textit{detail component}s  in different orders (\textit{i.e.}, \textit{k}=1, 2, 3, and 4). As seen, the experimental results are consistent with the above mathematical derivations.

Fig. \ref{fig_threeProperties} provides a toy example to demonstrate the  three properties of the \textit{detail component}, including separability, transferability, and  idempotence.
Firstly, the detail and shape components are separated from each model in the left column. Then, the detail component is crossly exchanged by  transferring to the other shape component in the middle column. Finally, 
the detail component is re-separated from the transferred surface, and used to compare with the source detail component to show the idempotence property in the right column, where {the MAE results} in the upper and lower error maps are 4.13$^\circ$ and 4.28$^\circ$, respectively.

\section{Application and Evaluation}
\label{sec:experiments}
Based on the properties of the proposed \textit{detail component}, three schemes are designed for surface geometry detail processing, including  \textit{geometry detail transfer}, \textit{geometric texture synthesis}, and \textit{3D surface super-resolution}. The significance of the proposed framework is that, by taking the \textit{detail component} as a feature carrier, the geometric surface texture is transformed to the intermediate feature map that can be processed as digital images.

\begin{figure*}[!t]
		\centering
		\begin{tabular}{p{3cm}<{\centering} p{3cm}<{\centering} p{3cm}<{\centering} p{3cm}<{\centering} p{3cm}<{\centering}}
			\includegraphics[height=3.3cm]{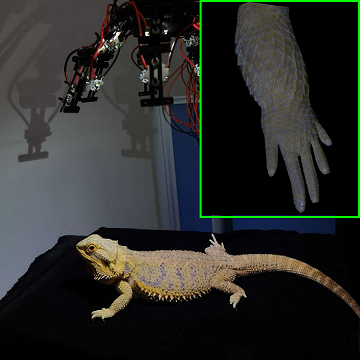}
			&\includegraphics[height=3.3cm]{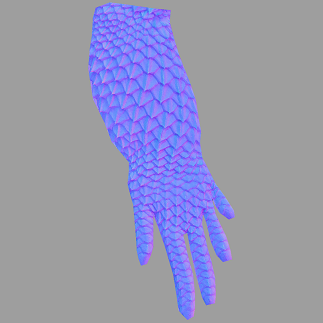}
			&\includegraphics[height=3.3cm]{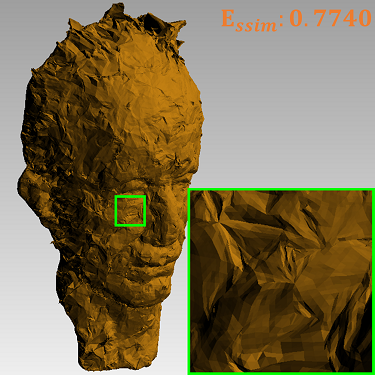}
			&\includegraphics[height=3.3cm]{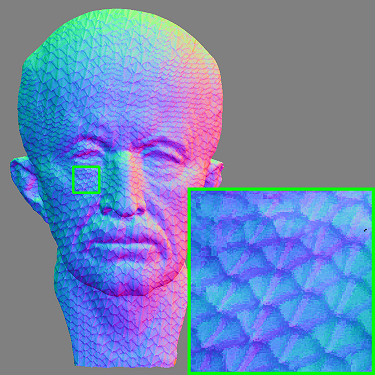}
			&\includegraphics[height=3.3cm]{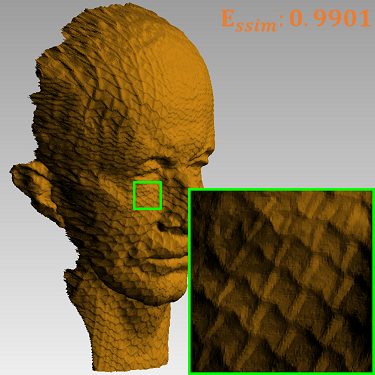}\\
			\small{\textit{Lizard}} & \small{\ \textit{Detail component}} & \small{{DGTS} \cite{hertz2020deep}} & \small{\textit{Synthesized normal}} &\small{\textit{Proposed}}\\
		\end{tabular} 
		\caption{\textbf{The \textit{geometric texture synthesis} result of the \textit{Lizard} model}. The \textit{Max Planck} model is selected as the target surface with a total of $1,665,036$ vertices. The \textit{detail component} feature map is captured and extracted as shown in the second figure (Best viewed by zooming in). }
		\label{fig_syn_lizard} 
\end{figure*}

\begin{figure*}[!t]
	\centering
	\begin{tabular}{p{3cm}<{\centering} p{3cm}<{\centering} p{3cm}<{\centering} p{3cm}<{\centering} p{3cm}<{\centering}}
		\includegraphics[height=3.3cm]{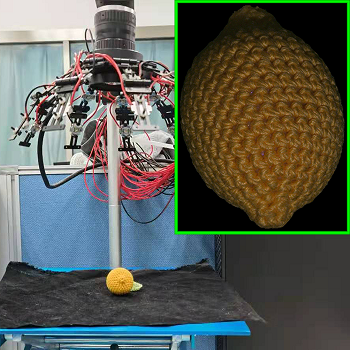}
		&\includegraphics[height=3.3cm]{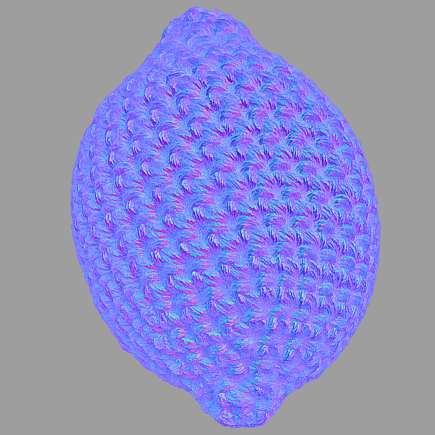}
		&\includegraphics[height=3.3cm]{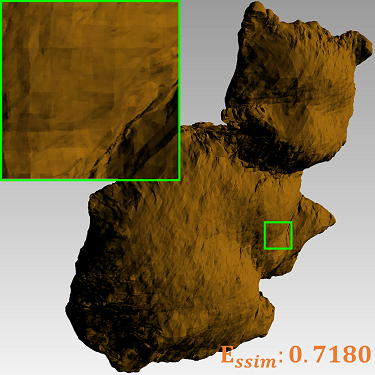}
		&\includegraphics[height=3.3cm]{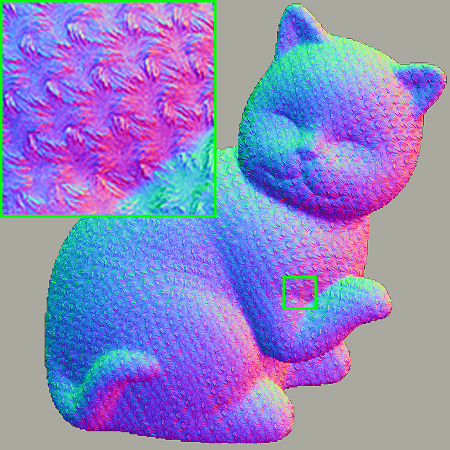}
		&\includegraphics[height=3.3cm]{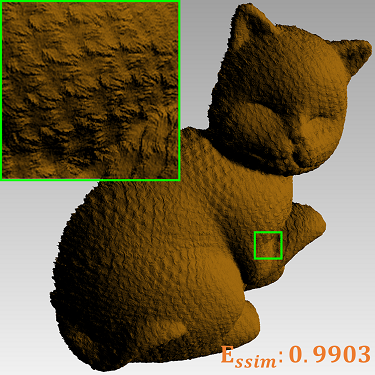}\\
		\small{\textit{Textile}} & \small{\ \textit{Detail component}} & \small{{DGTS} \cite{hertz2020deep}} & \small{\textit{Synthesized normal}} &\small{\textit{Proposed}}\\
	\end{tabular} 
	\caption{\textbf{The \textit{geometric texture synthesis} result of the \textit{Textile} model}. The \textit{Cat} model is selected as the target surface with a total of $1,999,807$ vertices. The $\mathbf{E}_{ssim}$ value of our proposed method is $0.9903$, while the $\mathbf{E}_{ssim}$ value of {DGTS} \cite{hertz2020deep} is $0.7180$.}
	\label{fig_syn_textile} 
\end{figure*}

\begin{figure*}[!t]
\centering
\begin{tabular}{p{1.4cm}<{\centering} p{1.4cm}<{\centering} p{1.4cm}<{\centering} p{1.4cm}<{\centering} p{1.4cm}<{\centering}p{1.4cm}<{\centering}p{1.4cm}<{\centering}p{1.4cm}<{\centering}p{1.4cm}<{\centering}p{1.4cm}<{\centering}}
\includegraphics[width=1.75cm]{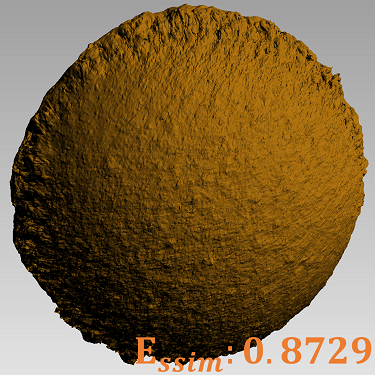}
&\includegraphics[width=1.75cm]{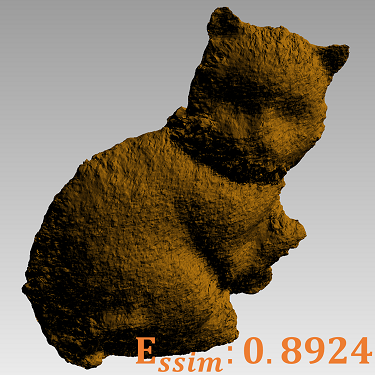}
&\includegraphics[width=1.75cm]{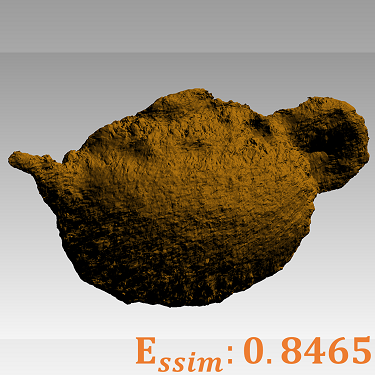}
&\includegraphics[width=1.75cm]{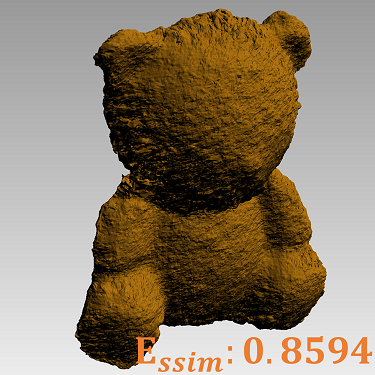}
&\includegraphics[width=1.75cm]{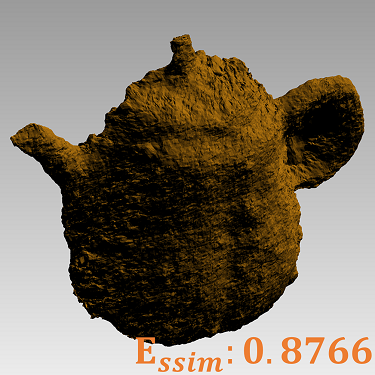}
&\includegraphics[width=1.75cm]{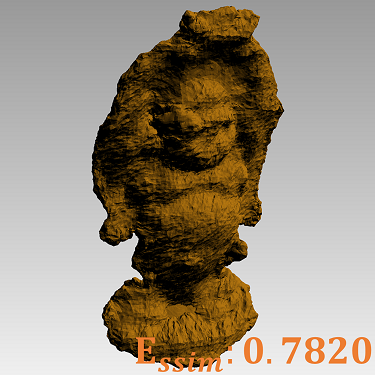}
&\includegraphics[width=1.75cm]{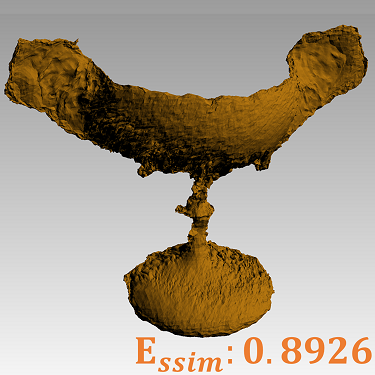}
&\includegraphics[width=1.75cm]{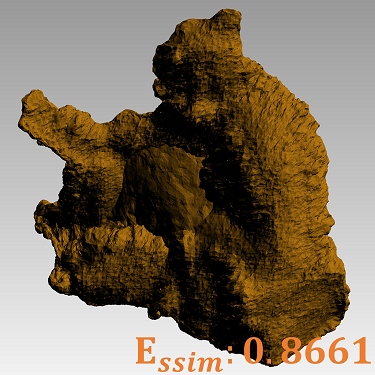}
&\includegraphics[width=1.75cm]{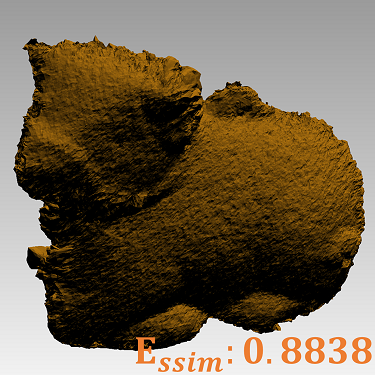}
&\includegraphics[width=1.75cm]{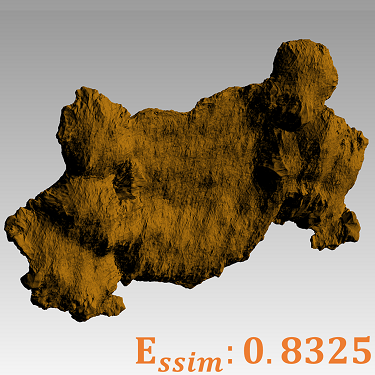}
\\
\includegraphics[width=1.75cm]{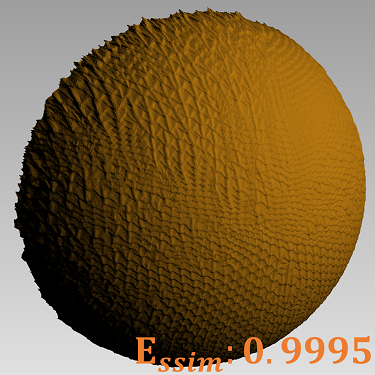}
&\includegraphics[width=1.75cm]{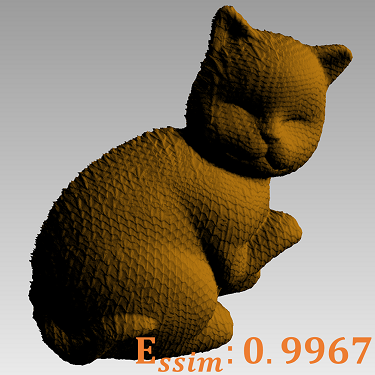}
&\includegraphics[width=1.75cm]{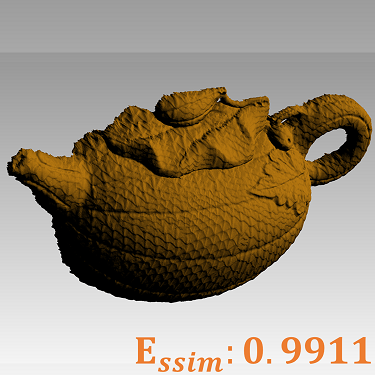}
&\includegraphics[width=1.75cm]{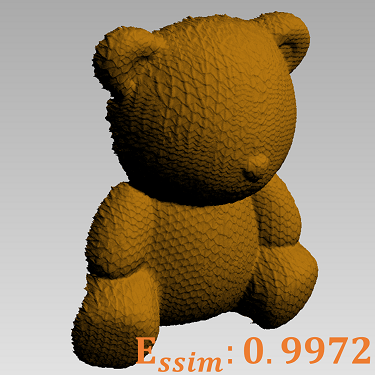}
&\includegraphics[width=1.75cm]{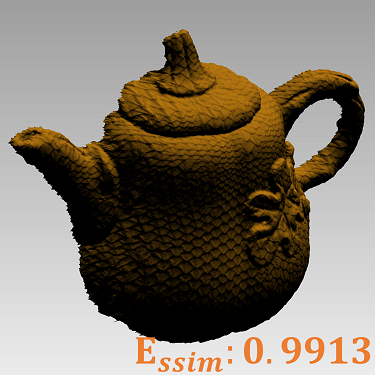}
&\includegraphics[width=1.75cm]{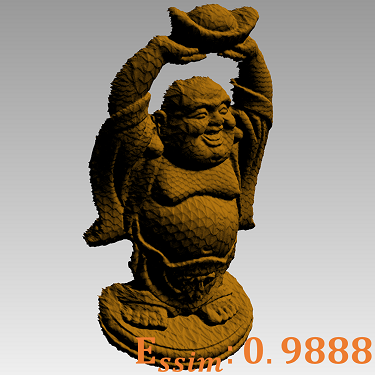}
&\includegraphics[width=1.75cm]{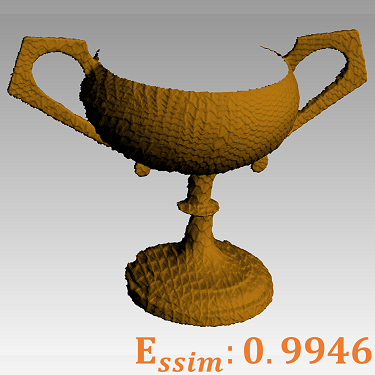}
&\includegraphics[width=1.75cm]{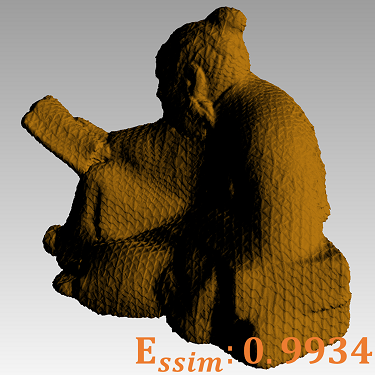}
&\includegraphics[width=1.75cm]{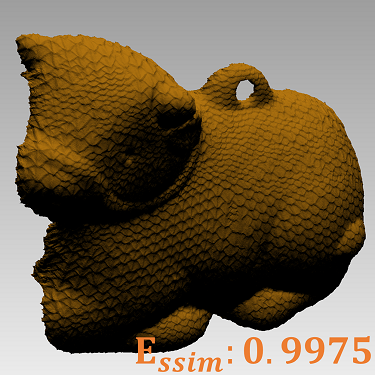}
&\includegraphics[width=1.75cm]{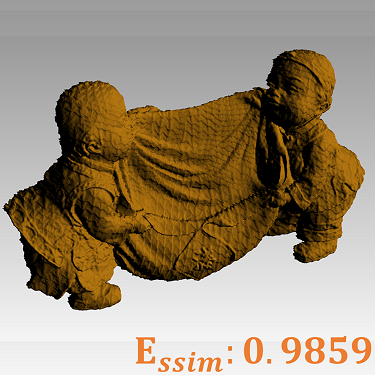}\\
\small{\textit{Ball}} & \small{\textit{Cat}} & \small{\textit{Pot1}} & \small{\textit{Bear}} & \small{\textit{Pot2}} &\small{\textit{Buddha}} &\small{\textit{Goblet}} & \small{\textit{Reading}} & \small{\textit{Cow}} & \small{\textit{Harvest}}\\
\end{tabular} 
\caption{\textbf{Comparative illustration of \textit{geometric texture synthesis} on the DiLiGenT dataset \cite{shi2019benchmark} using the \textit{Lizard} skin.} The top row are the texture synthesis results of {DGTS} \cite{hertz2020deep}, while the bottom row are the synthesis results of the proposed method.  From left to right: The $\mathbf{E}_{ssim}$ results of the \textit{detail component} in {DGTS} are provided at the bottom of each reconstructed object. Please zoom-in for details. }
\label{fig_syn_diligent} 
\end{figure*}

\subsection{Experimental Protocols}
\label{subsubsec:protocols}
\noindent \textbf{Input data}. The proposed framework does not limit the input type of data, as long as it can be converted into a single normal map.  It is noted that in this section, all 3D surfaces are reconstructed by the surface from normal (SfN) method \cite{xie2014surface}\footnote{https://charwill.github.io/SGP.html}.

For the input 3D data sources, they mainly come {from dense scanning devices}. The reason to select these data is due to the fact that they are more challenging and have not yet been properly processed in existing methods. There are two ways to obtain the real and dense data, \textit{i.e.}, by \textit{laser scanning} or \textit{photometric stereo} (PS). To this end, we built a PS device consisting of an industrial camera and $36$ LED lights (see Fig. \ref{fig_syn_lizard}) to obtain the real surface normal.

\noindent \textbf{Error measurement}. Considering that both the input and output of the proposed method are normals, it is reasonable to compare these results in the normal domain. Thus, the orientation difference between two compared normal vectors  is measured by $\mathbf{E}_{mae}$ \cite{xie2014surface,santo2017deep}. 
\begin{equation}\label{eq_mae_cal}
\small
\mathbf{E}_{mae}\left( {{\mathcal{N}_{o}},{\mathcal{N}_{t}}} \right) = \frac{1}{|\Omega|}\sum\limits_{\left( {r,c} \right) \in \Omega } {\arccos \left( {{\mathcal{N}_{o}}\left( {r,c} \right) \odot {\mathcal{N}_{t}}\left( {r,c} \right)} \right) },
\end{equation}
{where $|\Omega|$ denotes} the total number of all compared normal pairs in $\Omega$, ${\mathcal{N}_{o}}$ represents the reference input, and ${\mathcal{N}_{t}}$ represents the target output.  The $\mathbf{E}_{mae}$ value ranges from 0$^\circ$ to 180$^\circ$.

To comprehensively evaluate the generated surface quality, a widely-used structural similarity (SSIM) metric $\mathbf{E}_{ssim}$ in geometric surface processing  \cite{alldieck2019tex2shape,yuan2019face} is used to measure the difference between the original and resulted depths. 
$\mathbf{E}_{ssim}$ is calculated as the mean of the channel-wise SSIM value, ${\mathbf{E}_{ssim}}\left( {{\mathcal{N}_o},{\mathcal{N}_t}} \right) = \frac{1}{3}\sum\nolimits_{k} {E_{ssim}^k\left( {\mathcal{N}_o^k,\mathcal{N}_t^k} \right)}$, where the \textit{k}-th channel $\small{\mathbf{E}_{ssim}^k\left( {\mathcal{N}_o^k,\mathcal{N}_t^k} \right)}$ is defined as
\begin{equation}\label{eq_ssim_channelWise}
\resizebox{0.87\hsize}{!}{$\mathbf{E}_{ssim}^k\left( {\mathcal{N}_o^k,\mathcal{N}_t^k} \right) = 
        \frac{{\left( {2\mu _o^k\mu _t^k + c1} \right)\left( {\sigma _{ot}^k + c2} \right)}}{{\left( {{{\left( {\mu _o^k} \right)}^2} + {{\left( {\mu _t^k} \right)}^2} + c1} \right)\left( {{{\left( {\sigma _o^k} \right)}^2} + {{\left( {\sigma _t^k} \right)}^2} + c2} \right)}}
        $},
\end{equation}
where $\mu_o^k$ is the average of ${\mathcal{N}_o^k}$, $\mu_t^k$ is the average of ${\mathcal{N}_t^k}$, ${{{\left( {\sigma _o^k} \right)}^2}}$ is the variance of ${\mathcal{N}_o^k}$, ${{{\left( {\sigma _t^k} \right)}^2}}$ is the variance of ${\mathcal{N}_t^k}$, and ${\sigma _{ot}^k}$ is the covariance of ${\mathcal{N}_o^k}$ and ${\mathcal{N}_t^k}$. $c1$ and $c2$ are two constants used to maintain stability, where $c1$ = 0.01 and $c2$ = 0.03, respectively. The $\mathbf{E}_{ssim}$ value ranges from 0 to 1. When two compared images are identical, the $\mathbf{E}_{ssim}$ value is equal to one.
\begin{figure}[!t]
\raggedright
\begin{tabular}{p{4.0cm}<{\centering} p{4.0cm}<{\centering}}
\includegraphics[width=4.7cm]{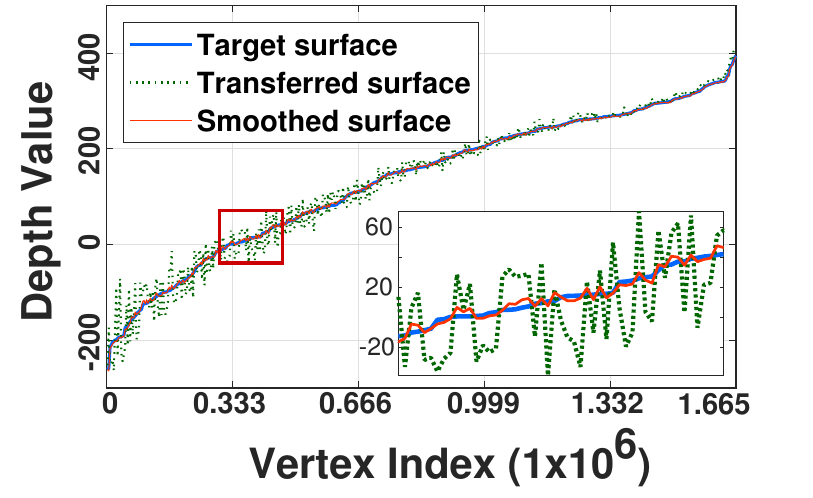}
&\includegraphics[width=4.7cm]{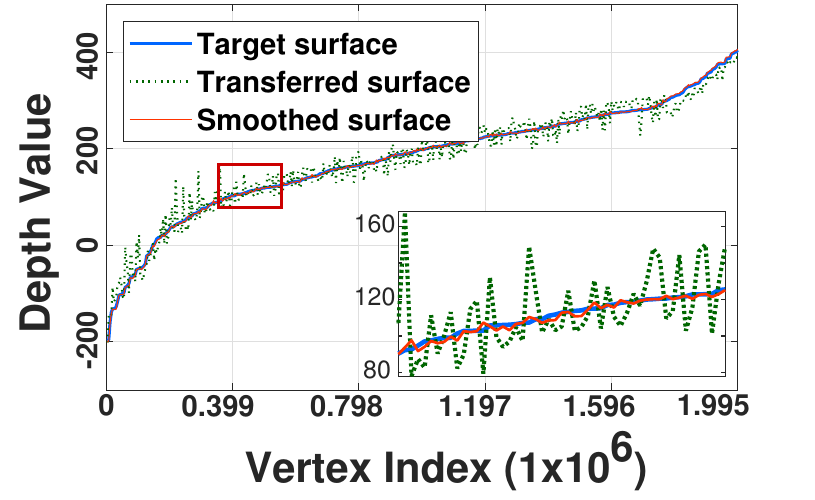}\\
\small{\textit{Lizard}} & \small{\textit{Textile}}\\
\end{tabular} 
\caption{\textbf{Performance evaluation of the proposed \textit{geometric texture synthesis} method in terms  of the height profile.} The blue solid line represents the original depth value, the green dot line represents the transferred depth value, and the red solid line represents the smoothed result of the transferred surface by a 5$\times$5 average filter.}
\label{fig_height_profile_gts} 
\end{figure}

\noindent \textbf{Computing platform}. All algorithms in the normal domain are implemented in MATLAB 2018, and performed on a computer with \textit{Intel i7-CPU@2.90GHz} and $32$GB \textit{RAM}. In addition, the CNN model is trained and tested on \textit{NVIDIA Tesla P100 GPU}.

In the experiments, many of the deep learning methods we have compared either do not provide training code or do not provide training databases. On the other hand, the related experiments are used to demonstrate that deep-learned models can be seamlessly embedded into the proposed framework. For a fair comparison, all the compared deep methods except RDN-Net \cite{zhang2018residual} directly adopt the off-the-shelf models provided by the authors.

In the experiments of \textit{3D surface super-resolution}, the training of RDN-Net \cite{zhang2018residual} is only trained on the original normal map, and that of ``Proposed'' denotes that \cite{zhang2018residual} is only trained on the \textit{detail component} map. The training samples are cropped into 192$\times$192 normal blocks, where the \textit{Adam} is used as our optimization solver in \textit{Python Toolbox PyTorch}. The initial learning rate is set to $10e$-$5$ and the batch size is 16. There are a total of $353$ samples used in training, testing, and validating the CNN models. The testing samples are strictly blind to the training process. All other training settings are the same as suggested \cite{zhang2018residual}.

\begin{table}[!t] 
		\centering 
		\caption{The MAE, SSIM, memory, and time of the transferred result between  \textit{Circular} and \textit{Bunny} with different resolutions (input points/vertices).} 
		\label{tab_transfer_circular}
		\begin{lrbox}{\tablebox}
			\renewcommand\arraystretch{1.2}
			\begin{tabular}{ccccccccc} 
				\hline\hline 
				\multicolumn{1}{c}{}
				&\multicolumn{1}{c}{Input points} 
				&\multicolumn{1}{c}{$\#5010$} & {$\#10106$} & {$\#20441$} & {$\#49532$} &{$\#153145$} &{$\#306491$}\\ 
				
				\hline
				\multicolumn{1}{c}{\multirow{3}{1.5cm}{Shape $\mathbf{E}_{mae}$(${}^{\circ}$)}}
				&{DSPL \cite{botsch2010polygon}}
				& {17.04} & {16.62} & {16.37} & {16.00} &n/a &n/a\\
				\multicolumn{1}{c}{}
				& {\cellcolor{mygray} LAPL \cite{sorkine2004laplacian}} & {\cellcolor{mygray}12.13} & {\cellcolor{mygray}11.07} & {\cellcolor{mygray}9.05} & {\cellcolor{mygray} n/a} & {\cellcolor{mygray} n/a} & {\cellcolor{mygray} n/a} \\
				\multicolumn{1}{c}{}
				&{\textbf{Proposed}}
				&{\textbf{4.97}} & {\textbf{4.16}} & {\textbf{3.48}} & {\textbf{2.79}} & {\textbf{2.42}} & {\textbf{2.28}}\\
				
				\hline
				\multicolumn{1}{c}{\multirow{3}{1.5cm}{Detail $\mathbf{E}_{mae}$(${}^{\circ}$)}}
				&{DSPL \cite{botsch2010polygon}}
				& {16.02} & {23.06} & {32.15} & {33.79} &n/a &n/a\\
				\multicolumn{1}{c}{}
				&{\cellcolor{mygray} LAPL \cite{sorkine2004laplacian}}
				& {\cellcolor{mygray}14.65} & {\cellcolor{mygray}17.11} & {\cellcolor{mygray}21.78} & {\cellcolor{mygray} n/a} & {\cellcolor{mygray} n/a} & {\cellcolor{mygray} n/a}\\
				\multicolumn{1}{c}{}
				&{\textbf{Proposed}}
				&{\textbf{10.71}} & {\textbf{10.93}} & {\textbf{10.87}} & {\textbf{10.90}} & {\textbf{10.86}} & {\textbf{10.86} } \\
				
				\hline
    			\multicolumn{1}{c}{\multirow{3}{1.5cm}{Detail $\mathbf{E}_{ssim}$}}
    			&{DSPL \cite{botsch2010polygon}}
    			& {0.92} & {0.83} & {0.74} & {0.72} & n/a & n/a\\
    			\multicolumn{1}{c}{}
    			& {\cellcolor{mygray} LAPL \cite{sorkine2004laplacian}}
    			& {\cellcolor{mygray} 0.93} & {\cellcolor{mygray} 0.90} & {\cellcolor{mygray} 0.86} &{\cellcolor{mygray} n/a} & {\cellcolor{mygray} n/a} & {\cellcolor{mygray} n/a} \\
    			\multicolumn{1}{c}{}
    			&{\textbf{Proposed}}
    			&{\textbf{0.9797}} & {\textbf{0.9809}} & {\textbf{0.9806}} & {\textbf{0.9805}} & {\textbf{0.9808}} &{\textbf{0.9808}} \\
\hline
				\multicolumn{1}{c}{\multirow{3}{1.5cm}{Memory (Mb)}}
				& {DSPL \cite{botsch2010polygon}}
				& {32.91} & {67.70} & {113.22} & {745.59} & n/a & n/a \\
				\multicolumn{1}{c}{}
				&{\cellcolor{mygray} LAPL \cite{sorkine2004laplacian}}
				& {\cellcolor{mygray} 827.52} & {\cellcolor{mygray} 3044.78} & {\cellcolor{mygray} 12620.90} & {\cellcolor{mygray} n/a} & {\cellcolor{mygray} n/a} & {\cellcolor{mygray} n/a} \\
				\multicolumn{1}{c}{}
				&{\textbf{Proposed}} & {\textbf{5.08}} & {\textbf{7.11}} & {\textbf{10.62}} &{\textbf{19.77}} &{\textbf{53.50}} &{\textbf{103.89}} \\
				
				\hline
				\multicolumn{1}{c}{\multirow{3}{1.5cm}{Time cost (Sec.)}}
				& {DSPL \cite{botsch2010polygon}}
				& {11.03} & {41.62} & {266.87} & {2365.20} & n/a & n/a  \\
				\multicolumn{1}{c}{}
				&{\cellcolor{mygray} LAPL \cite{sorkine2004laplacian}}
				& {\cellcolor{mygray} 20.93} & {\cellcolor{mygray} 89.64} & {\cellcolor{mygray} 383.56} & {\cellcolor{mygray} n/a} & {\cellcolor{mygray} n/a} & {\cellcolor{mygray} n/a} \\
				\multicolumn{1}{c}{}
				&{\textbf{Proposed}}
				&{\textbf{0.06}} & {\textbf{0.19}} & {\textbf{0.40}} &{\textbf{0.57}} & {\textbf{2.93}} &{\textbf{3.82}}\\
				\hline
				\hline  	
			\end{tabular}
\end{lrbox}
\scalebox{0.70}{\usebox{\tablebox}}	
\end{table}

\begin{figure}[!t]
\centering
\includegraphics[width=\linewidth]{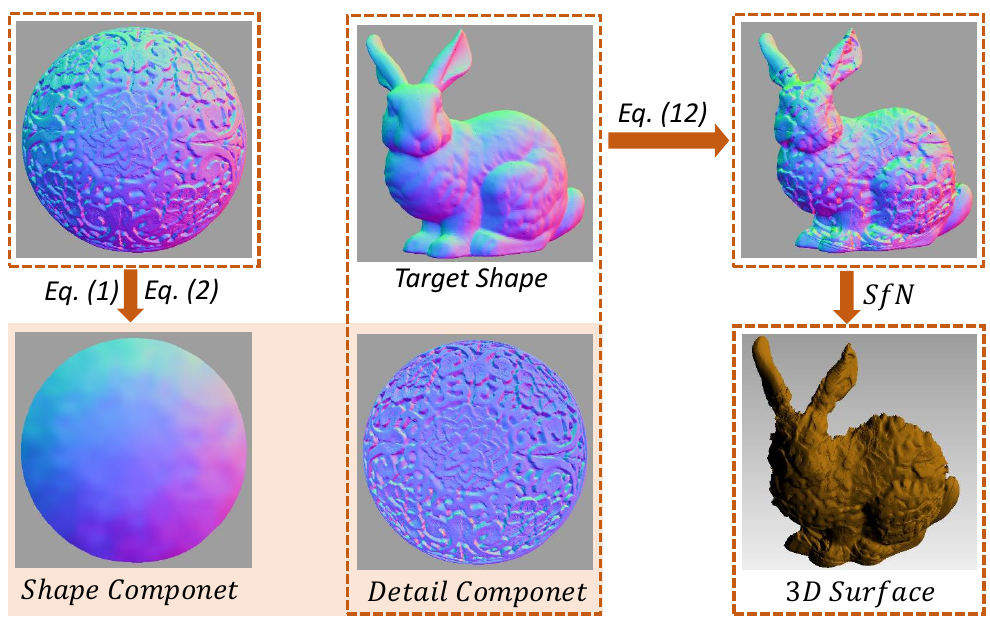}
\caption{\textbf{Flowchart of the \textit{geometry detail transfer}.} The \textit{shape component} and \textit{detail component} of the source normal are extracted by Eq. (\ref{eq_conv_cal}) and Eq. (\ref{eq_detail_form}), respectively.  Then, the \textit{detail component} is transferred to the target normal by Eq. (\ref{eq_transfer}) and reconstructed by the surface-from-normal (SfN) method \cite{xie2014surface}.}
\label{fig_transfer} 
\end{figure}

\subsection{Geometric Texture Synthesis}
\label{subsec:detailSynthesis}
In a practical texture synthesis, one common case is that the available source texture is not enough to be allocated on the target shape. One solution is to synthesize more similar textures from the source surface,  in order to cover the entire target surface shape. Unfortunately, existing geometry texture synthesis methods have not been well developed to deal with a dense and irregular texture pattern. In this section, the experiments are carried out by showing that the proposed \textit{detail component} can be used as {a standalone feature} map due to {its \textit{detail separability} property}, and hence it is possible to solve the problem of \textit{geometric texture synthesis}.

Specifically, for a small piece of {the source surface}, its \textit{detail component} is separated first, and then used to be synthesized in the feature map domain. In general, all the image texture synthesis methods are applicable to our proposed. For simplicity, we select a representative method \cite{efros1999texture} to demonstrate the {synthesis performance of the \textit{detail component}}. The synthesized result is then used to transfer on an entire target surface shape. The overall pipeline of the proposed \textit{geometric texture synthesis} scheme is illustrated in Fig. \ref{fig_synthesis}. 
\begin{figure*}[!t]
		\centering
		\begin{tabular}{p{2.5cm}<{\centering} p{2.5cm}<{\centering} p{2.5cm}<{\centering} p{2.5cm}<{\centering} p{2.5cm}<{\centering} p{2.5cm}<{\centering}}
			\includegraphics[height=2.8cm]{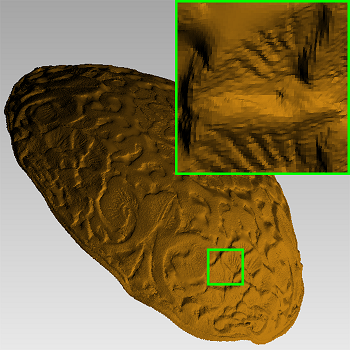}
			&\includegraphics[height=2.8cm]{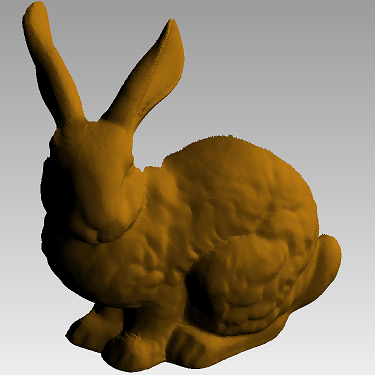}
			&\includegraphics[height=2.8cm]{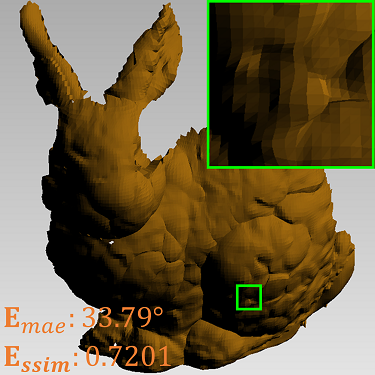}
			&\includegraphics[height=2.8cm]{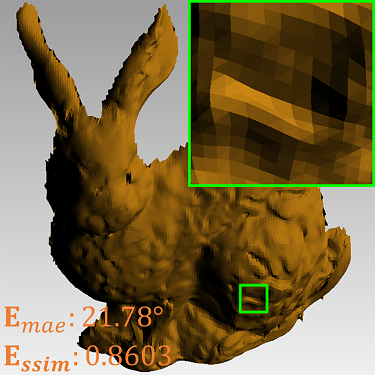}
			&\includegraphics[height=2.8cm]{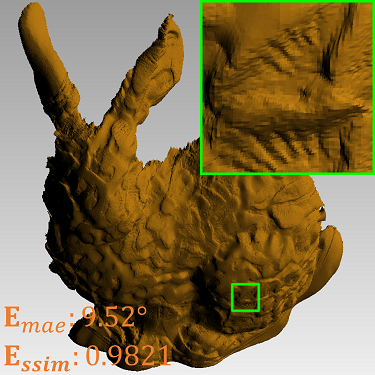}
            &\includegraphics[height=2.8cm]{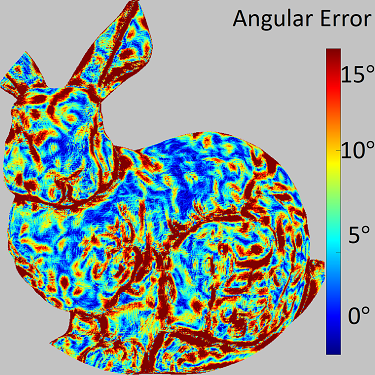}\\
			\small{\textit{Original surface}}  & \small{\textit{Target surface}} & \small{DSPL \cite{botsch2010polygon}} & \small{LAPL \cite{sorkine2004laplacian}} &\small{\textit{Proposed}} &\small{{Energy map}}\\
		\end{tabular} 
		\caption{\textbf{Illustration of the \textit{geometry detail transfer} for the \textit{Bunny} model}. The original surface is \textit{Circular} with the total vertex number of $646,693$. The \textit{ detail component} feature map used for transfer is reconstructed and shown in the right-top green box. {The energy map of angular error is measured between the original \textit{detail component} and our transferred one, and $\mathbf{E}_{mae}$ is $9.52^\circ$.}}
		\label{fig_transfer_full} 		
		
\end{figure*}

Fig. \ref{fig_syn_lizard} shows a lizard, which is reconstructed by our   
\textit{photometric stereo} setup. It is noted that only a part of the scanned skin normal data is used for synthesis, and is named as the \textit{Lizard} model. One latest deep-learned geometry synthesis method  {DGTS} \cite{hertz2020deep} is used for comparison. As seen in Fig. \ref{fig_syn_lizard}, {DGTS} cannot properly synthesize the dense and non-repetitive texture pattern, where the $\mathbf{E}_{ssim}$ value is only $0.7740$.  The main reason is that its ability to capture the texture feature depends on the receptive field. In addition, it also has the disadvantage of incurring heavy time cost and excessive memory cost,  \textit{e.g.},  $233$Sec and $14,216$Mb. In contrast, the transferred result of our method has the $\mathbf{E}_{ssim}$ value of $0.9901$, and the running time and memory costs are $11.88$Sec and $857$Mb, respectively.

Fig. \ref{fig_syn_textile} shows another synthesis example with complex textures, where the reconstructed  \textit{Textile} model is used as {the source \textit{detail component}}, and the \textit{Cat} model is used as the target shape. While {{DGTS} \cite{hertz2020deep} produces a poor \textit{geometric texture synthesis} result, where the related $\mathbf{E}_{ssim}$ value of the \textit{detailed component} between the original and {DGTS} is 0.7180, and our method generates a satisfying texture synthesis result, where the related $\mathbf{E}_{ssim}$ value is 0.9903.}

Fig. \ref{fig_syn_diligent} provides {the overall visual results of \textit{geometric texture synthesis} on the DiLiGenT dataset \cite{shi2019benchmark}}. As seen, our proposed texture synthesis scheme can preserve the geometric details of the original surface as well as the surface shape of the target 3D model. {{DGTS} \cite{hertz2020deep} produces obviously worse synthesized results than the proposed method. At the same time, {DGTS} also deteriorates the target shape. For example, the face of \textit{Buddha} by {DGTS} is difficult to be perceived.}

\begin{figure*}[!t]
\centering
		\begin{tabular}{p{2.5cm}<{\centering} p{2.5cm}<{\centering} p{2.5cm}<{\centering} p{2.5cm}<{\centering} p{2.5cm}<{\centering} p{2.5cm}<{\centering}}
			\includegraphics[height=2.8cm,width=2.8cm]{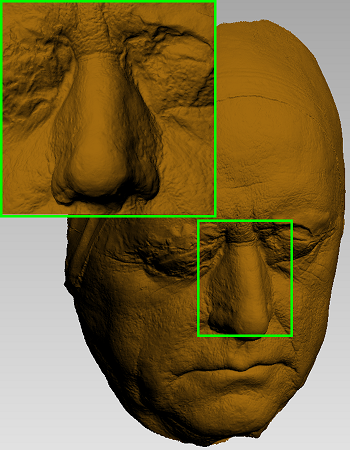}
			&\includegraphics[height=2.8cm,width=2.8cm]{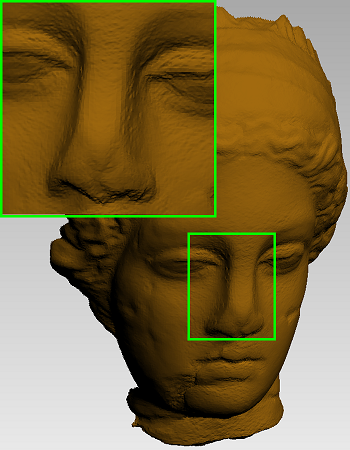}
			&\includegraphics[height=2.8cm,width=2.8cm]{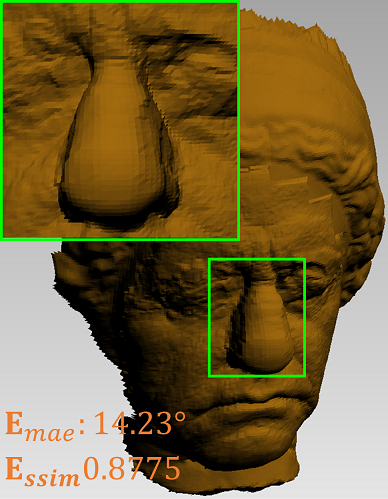}
			&\includegraphics[height=2.8cm,width=2.8cm]{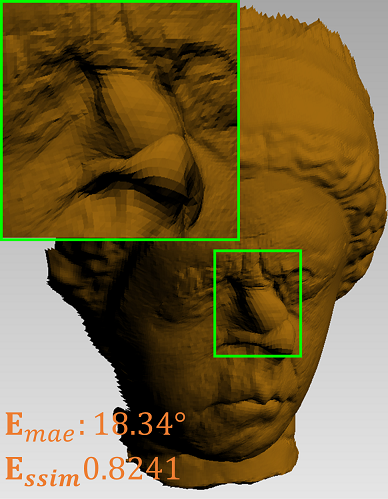}
			&\includegraphics[height=2.8cm,width=2.8cm]{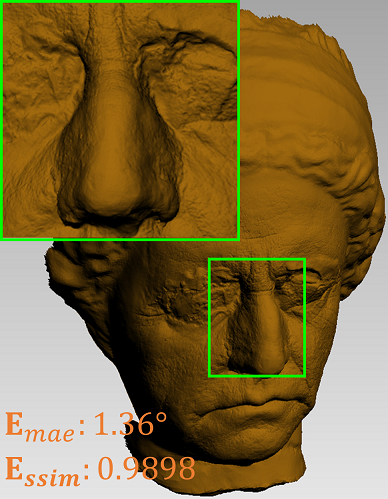}
			&\includegraphics[height=2.8cm,width=2.8cm]{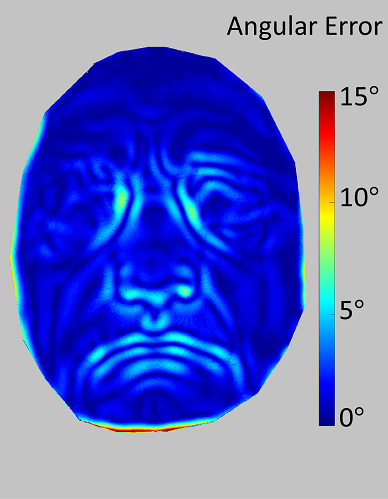}\\
			\small{\textit{Original surface}}  & \small{\textit{Target surface}} & \small{DSPL \cite{botsch2010polygon}} & \small{LAPL  \cite{sorkine2004laplacian}} &\small{\textit{Proposed}} &\small{{Energy map}}\\
		\end{tabular} 
		\caption{\textbf{Performance comparisons of the local transfer.} The local face of \textit{Venus} is replaced by \textit{Goethe}, where the source \textit{Goethe} has the total vertex number of $539,352$.  A partial zoom-in part is shown in the left-top green box. {The energy map of angular error is measured between the original \textit{detail component} and  our transferred one, and  $\mathbf{E}_{mae}$ is $1.36^\circ$.}}
\label{fig_transfer_local_venus} 
\end{figure*}

\begin{figure*}[!t]
	\centering
		\begin{tabular}{p{2.5cm}<{\centering} p{2.5cm}<{\centering} p{2.5cm}<{\centering} p{2.5cm}<{\centering} p{2.5cm}<{\centering} p{2.5cm}<{\centering}}
		\includegraphics[height=2.8cm]{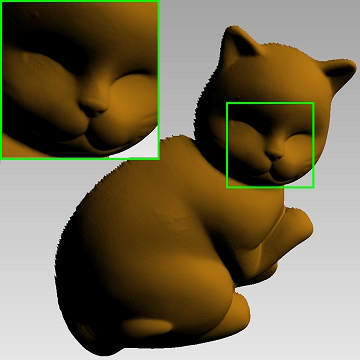}
		&\includegraphics[height=2.8cm]{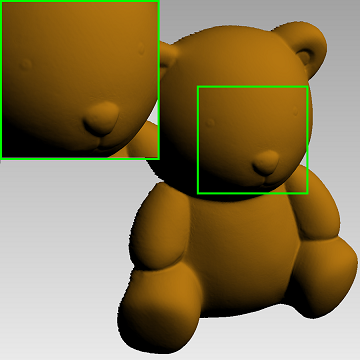}
		&\includegraphics[height=2.8cm]{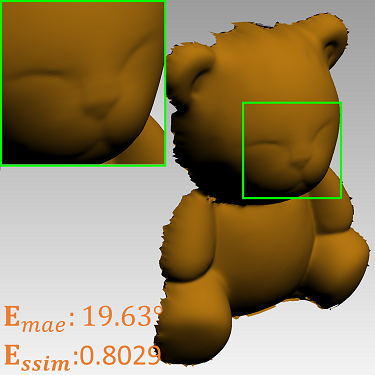}
		&\includegraphics[height=2.8cm]{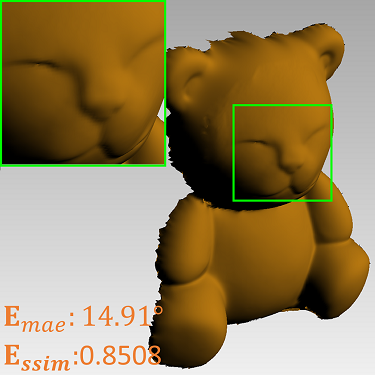}
		&\includegraphics[height=2.8cm]{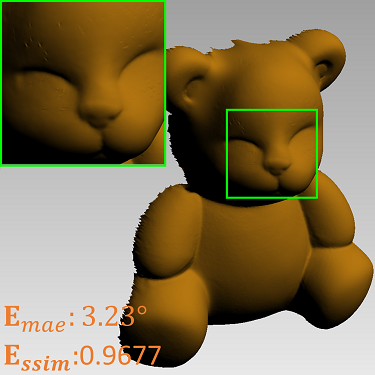}
		&\includegraphics[height=2.8cm]{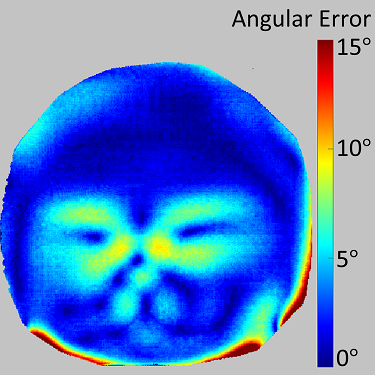}\\
		\small{\textit{Original surface}}  & \small{\textit{Target surface}} & \small{DSPL \cite{botsch2010polygon}} & \small{LAPL \cite{sorkine2004laplacian}} &\small{\textit{Proposed}} &\small{{Energy map}}\\
	\end{tabular} 
	\caption{\textbf{Performance comparisons of the local transfer.} The local face of \textit{Bear} is replaced by \textit{Cat}, where the source \textit{Cat} has the total vertex number of $119,389$. A zoom-in part is shown in the left-top green box. {The energy map of angular error is measured between the original \textit{detail component} and our transferred one, and $\mathbf{E}_{mae}$ is $3.23^\circ$.}}
	\label{fig_transfer_local_bear} 
\end{figure*}

\begin{figure*}[!t]
	\centering
		\begin{tabular}{p{2.5cm}<{\centering} p{2.5cm}<{\centering} p{2.5cm}<{\centering} p{2.5cm}<{\centering} p{2.5cm}<{\centering} p{2.5cm}<{\centering}}
		\includegraphics[height=2.8cm]{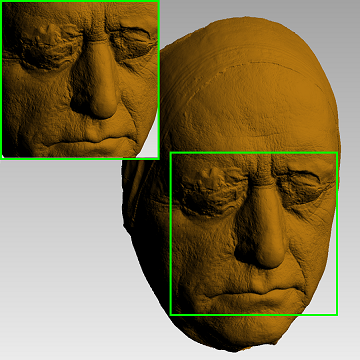}
		&\includegraphics[height=2.8cm]{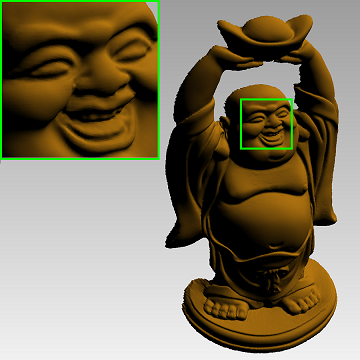}
		&\includegraphics[height=2.8cm]{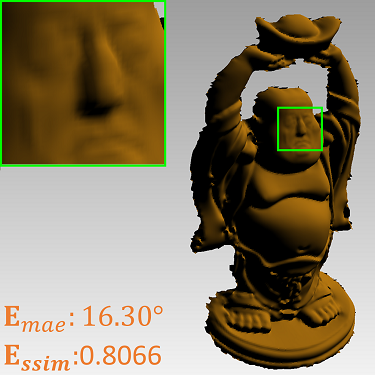}
		&\includegraphics[height=2.8cm]{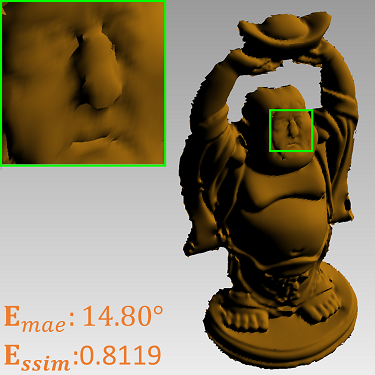}
		&\includegraphics[height=2.8cm]{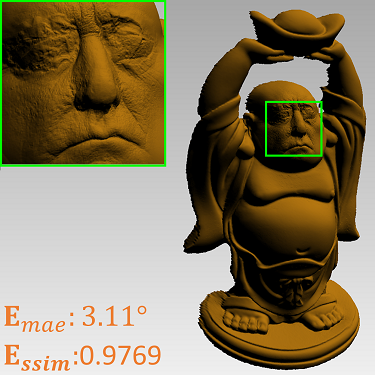}
		&\includegraphics[height=2.8cm]{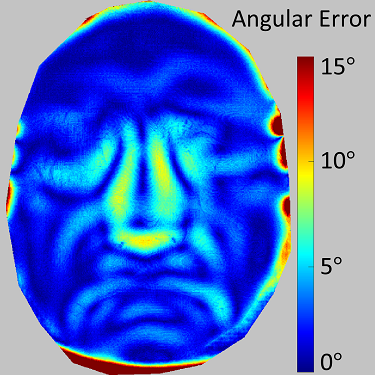}\\
		\small{\textit{Original surface}}  & \small{\textit{Target surface}} & \small{DSPL \cite{botsch2010polygon}} & \small{LAPL \cite{sorkine2004laplacian}} &\small{\textit{Proposed}} &\small{{Energy map}}\\
	\end{tabular} 
	\caption{\textbf{Performance comparisons of the local transfer.} The local face of \textit{Buddha} is replaced by \textit{Goethe}, where the source \textit{Goethe} has the total vertex number of $334,696$. A zoom-in part is shown in the left-top green box.  {The energy map of angular error is measured between the original \textit{detail component} and our transferred one, and $\mathbf{E}_{mae}$ is $3.11^\circ$.}}
	\label{fig_transfer_local_buddha} 
\end{figure*}

Moreover, we evaluate the transferred surface using the height profile as shown in Fig. \ref{fig_height_profile_gts}, where the depth values of a    target surface are firstly reordered in an ascending order, and the related depth values of a transferred surface are plotted according to  their positions. The depth values of the proposed method are similar to the target surface, which also validates the shape similarity in the property of the \textit{detail transferability}.

\subsection{Geometry Detail Transfer}
For \textit{geometry detail transfer}, the \textit{detail component} and \textit{shape component} of {the source and target} surfaces are separated by Eq. (\ref{eq_conv_cal}) and Eq. (\ref{eq_detail_form}), respectively. The proposed detail transfer scheme is illustrated in Fig. \ref{fig_transfer}. The \textit{detail component} of the original surface is transferred to the \textit{shape component} of the target surface.
In this section, the performance of the \textit{geometry detail transfer} is evaluated and compared with two mesh-based surface editing methods.

The whole {source surface} is transferred to a target shape, which is marked as global transfer. Fig. \ref{fig_transfer_full} shows that \textit{Bunny} is transferred with the geometry details of \textit{Circular}. The \textit{Circular} model is a high-definition laser-scanning data downloaded from \textit{Aim@Shape}, and its normal map with resolution of $1000 \times 1000$ are obtained by \textit{MeshLab} as the input. Two mesh-based methods, \textit{i.e.}, the Laplace mesh editing method (denoted as LAPL) \cite{sorkine2004laplacian} and the normal displacement method (denoted as DSPL) \cite{botsch2010polygon}, are used for comparisons. Compared with the proposed method,  neither of them can properly deal with the real dense surface. In addition, the transferred surface is evaluated in both the shape and detail aspects. For surface shape comparison, the $\mathbf{E}_{mae}$ error between the \textit{shape component} of \textit{Bunny} and that of the transferred result is calculated. For geometry details comparison, the $\mathbf{E}_{mae}$ and $\mathbf{E}_{ssim}$ results between the \textit{detail component} of \textit{Circular} and that of the transferred result are also obtained. Table \ref{tab_transfer_circular} provides the detailed MAE, SSIM, memory, and {running time of the transfer process} between  \textit{Circular} and \textit{Bunny} with various resolutions (input points/vertices), where  ``n/a'' denotes that the results can not be obtained on the current computing platform. 
For instance, under the same amount of the input vertices, like $5,000$, LAPL requires more than $800$Mb memory, which is $162$ times of ours. What makes the matter worse is that the memory gap increases exponentially with the number of input vertices. 
As for the time cost comparison, our method can calculate the dense surface in a trivial amount of time compared with DSPL and  LAPL.

\begin{figure*}[!t]
\centering
\begin{tabular}{p{2.5cm}<{\centering} p{2.5cm}<{\centering} p{2.5cm}<{\centering} p{2.5cm}<{\centering} p{2.5cm}<{\centering} p{2.5cm}<{\centering}}
\includegraphics[height=2.8cm]{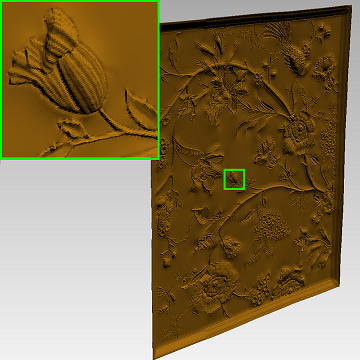}
&\includegraphics[height=2.8cm]{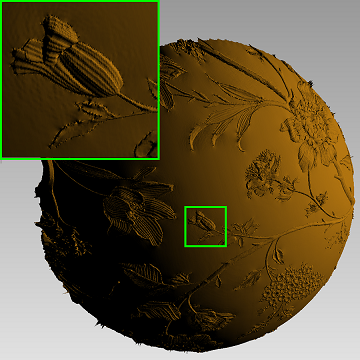}
&\includegraphics[height=2.8cm]{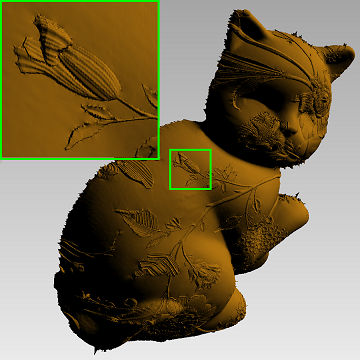}
&\includegraphics[height=2.8cm]{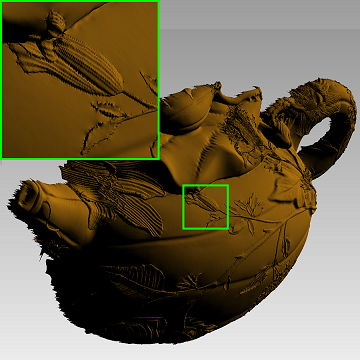}
&\includegraphics[height=2.8cm]{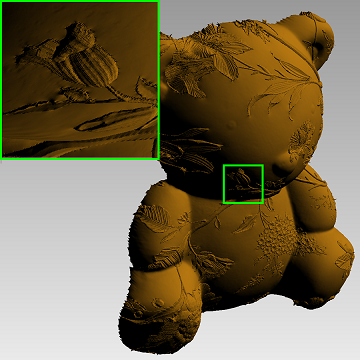}
&\includegraphics[height=2.8cm]{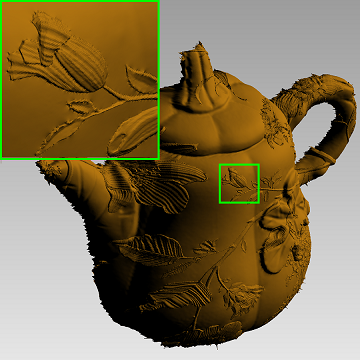}\\
\small{\textit{Original surface}}  &\small{\textit{Ball}} & \small{\textit{Cat}} & \small{\textit{Pot1}} & \small{\textit{Bear}} & \small{\textit{Pot2}}\\
\includegraphics[width=2.8cm,height=2.8cm]{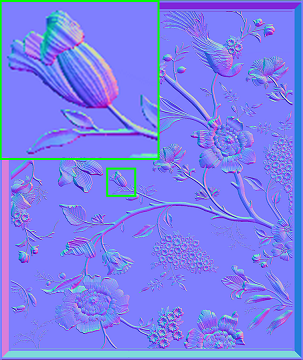}
&\includegraphics[height=2.8cm]{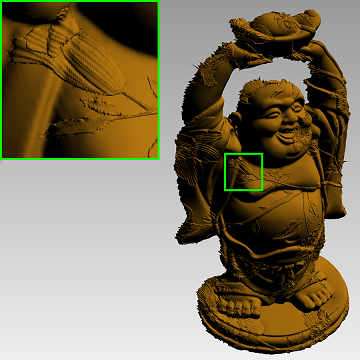}
&\includegraphics[height=2.8cm]{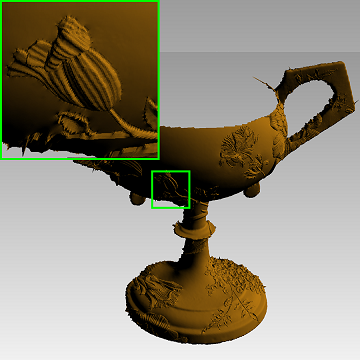}
&\includegraphics[height=2.8cm]{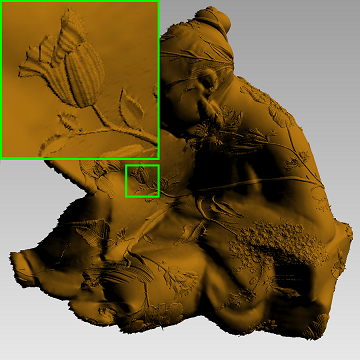}
&\includegraphics[height=2.8cm]{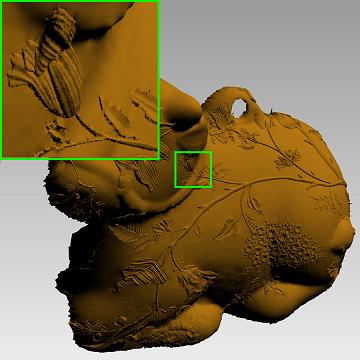}
&\includegraphics[height=2.8cm]{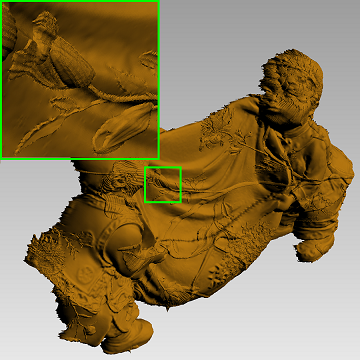}\\
\small{\textit{Normal}}  &\small{\textit{Buddha}} &\small{\textit{Goblet}} & \small{\textit{Reading}} & \small{\textit{Cow}} & \small{\textit{Harvest}}\\
\end{tabular} 
\caption{\textbf{Performance comparisons of the global transfer.} The original model \textit{Flower} is transferred  on the DiLiGenT dataset  \cite{shi2019benchmark}. A zoom-in part is shown in the left-top green box.}
\label{fig_transfer_diligent} 
\end{figure*}

\begin{table*}[!t] 
\centering 
\caption{Quantitative results of the \textit{geometric detail transfer} on the latest DiLiGenT dataset  \cite{shi2019benchmark}. } 
\label{tab:gdt_results}
\begin{lrbox}{\tablebox}
\setlength{\tabcolsep}{3.8mm}
\renewcommand\arraystretch{1.2}
\begin{tabular}{cccccccccccc} 
\hline\hline 
\multicolumn{1}{c}{}
&\multicolumn{1}{c}{Method} 
			&\multicolumn{1}{c}{\textit{Ball}} & {\textit{Cat}} & {\textit{Pot1}} & {\textit{Bear}} & {\textit{Pot2}} & {\textit{Buddha}} & {\textit{Goblet}} & {\textit{Reading}} & {\textit{Cow}} & {\textit{Harvest}}\\ 
			
			\hline
			\multicolumn{1}{c}{\multirow{3}{1.5cm}{Shape $\mathbf{E}_{mae}$(${}^{\circ}$)}}
			&{DSPL \cite{botsch2010polygon}}
			& {4.10} & {5.93} & {8.83} & {7.13} & {3.18} & {4.90} & {4.82} & {9.43} & {4.04} & {2.51}\\
			\multicolumn{1}{c}{}
			& {\cellcolor{mygray} LAPL \cite{sorkine2004laplacian}} & {\cellcolor{mygray} 1.73} & {\cellcolor{mygray} 2.22} & {\cellcolor{mygray} 2.85} & {\cellcolor{mygray} 3.17} & {\cellcolor{mygray} 4.37} & {\cellcolor{mygray} 6.49} & {\cellcolor{mygray} 3.39}
			& {\cellcolor{mygray} 3.02} & {\cellcolor{mygray} 2.25} & {\cellcolor{mygray} 6.10}\\
			\multicolumn{1}{c}{}
			&{\textbf{Proposed}}
			&{\textbf{1.83}} & {\textbf{2.02}} & {\textbf{2.49}} & {\textbf{1.96}} & {\textbf{2.26}} & {\textbf{2.54}} & {\textbf{2.90}} & {\textbf{2.09}} & {\textbf{1.90}} & {\textbf{2.43}}\\
			
			\hline
			\multicolumn{1}{c}{\multirow{3}{1.5cm}{Shape $\mathbf{E}_{ssim}$}}
			&{DSPL \cite{botsch2010polygon}}
			& {0.9895} & {0.9944} & {0.9868} & {0.9827} & {0.9979} & {0.9950} & {0.9972} & {0.9744} & {0.9945} & {0.9992}\\
			\multicolumn{1}{c}{}
			& {\cellcolor{mygray} LAPL \cite{sorkine2004laplacian}} & {\cellcolor{mygray} 0.9983} & {\cellcolor{mygray} 0.9990} & {\cellcolor{mygray} 0.9984} & {\cellcolor{mygray} 0.9956} & {\cellcolor{mygray} 0.9959} & {\cellcolor{mygray} 0.9916} & {\cellcolor{mygray} 0.9987}
			& {\cellcolor{mygray} 0.9979} & {\cellcolor{mygray} 0.9988} & {\cellcolor{mygray} 0.9946}\\
			\multicolumn{1}{c}{}
			&{\textbf{Proposed}}
			&{\textbf{0.9990}} & {\textbf{0.9995}} & {\textbf{0.9995}} & {\textbf{0.9994}} & {\textbf{0.9995}} & {\textbf{0.9994}} & {\textbf{0.9994}} & {\textbf{0.9994}} & {\textbf{0.9995}} & {\textbf{0.9994}}\\
			
			\hline
			\multicolumn{1}{c}{\multirow{3}{1.5cm}{Detail $\mathbf{E}_{mae}$(${}^{\circ}$)}}
			&{DSPL \cite{botsch2010polygon}}
			& {16.15} & {19.67} & {24.31} & {19.91} & {19.57} & {26.17} & {24.63} & {23.73} & {16.90} & {20.50}\\
			\multicolumn{1}{c}{}
			& {\cellcolor{mygray} LAPL \cite{sorkine2004laplacian}} & {\cellcolor{mygray} 13.90} & {\cellcolor{mygray} 17.53} & {\cellcolor{mygray} 18.73} & {\cellcolor{mygray} 15.19} & {\cellcolor{mygray} 18.14} & {\cellcolor{mygray} 24.71} & {\cellcolor{mygray} 20.03}
			& {\cellcolor{mygray} 18.03} & {\cellcolor{mygray} 15.13} & {\cellcolor{mygray} 21.73}\\
			\multicolumn{1}{c}{}
			&{\textbf{Proposed}}
			&{\textbf{1.43}} & {\textbf{4.13}} & {\textbf{6.83}} & {\textbf{3.55}} & {\textbf{6.17}} & {\textbf{13.02}} & {\textbf{5.69}} & {\textbf{5.67}} & {\textbf{3.57}} & {\textbf{10.63}}\\
			
			\hline
			\multicolumn{1}{c}{\multirow{3}{1.5cm}{Detail $\mathbf{E}_{ssim}$}}
			&{DSPL \cite{botsch2010polygon}}
			& {0.9088} & {0.9205} & {0.9307} & {0.8959} & {0.9443} & {0.9158} & {0.9531} & {0.8826} & {0.9265} & {0.9291}\\
			\multicolumn{1}{c}{}
			& {\cellcolor{mygray} LAPL \cite{sorkine2004laplacian}} & {\cellcolor{mygray} 0.9385} & {\cellcolor{mygray} 0.9445} & {\cellcolor{mygray} 0.9603} & {\cellcolor{mygray} 0.9438} & {\cellcolor{mygray} 0.9491} & {\cellcolor{mygray} 0.9269} & {\cellcolor{mygray} 0.9665}
			& {\cellcolor{mygray} 0.9369} & {\cellcolor{mygray} 0.9513} & {\cellcolor{mygray} 0.9251}\\
			\multicolumn{1}{c}{}
			&{\textbf{Proposed}}
			&{\textbf{0.9994}} & {\textbf{0.9964}} & {\textbf{0.9933}} & {\textbf{0.9966}} & {\textbf{0.9933}} & {\textbf{0.9805}} & {\textbf{0.9966}} & {\textbf{0.9933}} & {\textbf{0.9971}} & {\textbf{0.9848}}\\
			\hline
			
			\hline
			\multicolumn{1}{c}{\multirow{3}{1.5cm}{Memory (Mb)}}
			&{DSPL \cite{botsch2010polygon}}
			& {305.24} & {305.23} & {310.66} & {301.87} & {304.72} & {308.43} & {303.31} & {311.93} & {304.59} & {311.30}\\
			\multicolumn{1}{c}{}
			& {\cellcolor{mygray} LAPL \cite{sorkine2004laplacian}} & {\cellcolor{mygray} 12183.22} & {\cellcolor{mygray} 11616.61} & {\cellcolor{mygray} 11694.25} & {\cellcolor{mygray} 11587.38} & {\cellcolor{mygray} 12276.66} & {\cellcolor{mygray} 11988.65} & {\cellcolor{mygray} 12760.16}
			& {\cellcolor{mygray} 11457.79} & {\cellcolor{mygray} 11853.22} & {\cellcolor{mygray} 10910.99}\\
			\multicolumn{1}{c}{}
			&{\textbf{Proposed}}
			&{\textbf{17.66}} & {\textbf{18.75}} & {\textbf{21.76}} & {\textbf{17.64}} & {\textbf{19.91}} & {\textbf{19.23}} & {\textbf{25.82}} & {\textbf{17.63}} & {\textbf{17.39}} & {\textbf{20.10}}\\
				
			\hline
			\multicolumn{1}{c}{\multirow{3}{1.5cm}{Time cost (sec.)}}
			&{DSPL \cite{botsch2010polygon}}
			& {5.26} & {5.05} & {4.88} & {4.60} & {4.64} & {4.87} & {4.61} & {4.83} & {4.77} & {4.44}\\
			\multicolumn{1}{c}{}
			& {\cellcolor{mygray} LAPL \cite{sorkine2004laplacian}} & {\cellcolor{mygray} 696.67} & {\cellcolor{mygray} 658.30} & {\cellcolor{mygray} 688.34} & {\cellcolor{mygray} 586.96} & {\cellcolor{mygray} 675.27} & {\cellcolor{mygray} 819.87} & {\cellcolor{mygray} 676.72} & {\cellcolor{mygray} 706.23} & {\cellcolor{mygray} 647.91} & {\cellcolor{mygray} 733.78}\\
			\multicolumn{1}{c}{}
			&{\textbf{Proposed}}
			&{\textbf{0.50}} & {\textbf{0.48}} & {\textbf{0.52}} & {\textbf{0.40}} & {\textbf{0.44}} & {\textbf{0.50}} & {\textbf{0.59}} & {\textbf{0.42}} & {\textbf{0.47}} & {\textbf{0.50}}\\
			\hline
			\hline  
			
\end{tabular}
\end{lrbox}
\scalebox{0.75}{\usebox{\tablebox}}	
\end{table*}

Different from the global transfer, local transfer means that {a local geometry feature} is transferred to the target surface, which can be regarded as coarse-grained detail transfer. It is worth noting that the detail granularity depends on that, the closer the \textit{shape component} in Eq. \eqref{eq_conv_cal} is to the original surface, the finer the granularity of the \textit{detail component} in Eq. \eqref{eq_detail_form}, and vice versa.  Fig. \ref{fig_transfer_local_venus} shows an interesting example of the local transfer, where the face of \textit{Venus} (downloaded from \textit{Aim@Shape}) zoomed in the green bounding box is replaced by \textit{Goethe}.  \textit{Goethe} is used to extract the coarse-grained detail by setting a large filter size ($w$=10) in Eq. \eqref{eq_conv_cal} to get {the coarse geometry feature}. The transferred results show that LAPL \cite{sorkine2004laplacian} and DSPL \cite{botsch2010polygon} cannot preserve the nose shape of \textit{Goethe}. Meanwhile, {the proposed method generates} the best visual effect, where the transferred nose shape is almost the same as the original one. The transferred face region of our method has the $\mathbf{E}_{mae}$ value of $1.36^\circ$ and the $\mathbf{E}_{ssim}$ value of $0.9898$, respectively.	Figs. \ref{fig_transfer_local_bear} and \ref{fig_transfer_local_buddha} show another two examples of the local surface transfer. The transferred faces of \textit{Cat} and \textit{Goethe} are satisfactorily matched with the {target shapes, which further demonstrate} the effectiveness of the proposed normal-based detail representation.

In addition, Table \ref{tab:gdt_results} provides the quantitative  evaluation results for the \textit{geometric detail transfer} on the DiLiGenT dataset \cite{shi2019benchmark}, and the {relevant visual results are provided in Fig. \ref{fig_transfer_diligent}.  We  provide the} detailed results by transferring  \textit{Flowers} to ten different target surfaces. The average results of the shape MAE for DSPL, LAPL and our method are  5.48$^\circ$, 3.56$^\circ$, and 2.24$^\circ$, respectively. Meanwhile, the average results of the detail MAE for DSPL, LAPL, and our method are  21.15$^\circ$, 18.31$^\circ$, and 6.07$^\circ$, respectively.  It shows that our method can greatly improve the shape and detail transfer performance compared with DSPL and LAPL. In addition, we run the same experiments for ten times to collect the average of memory cost (Mb) and time cost (Sec) to measure the efficiency of each scheme. While the average results of the memory consumption for DSPL, LAPL and our method are 306.73, 11833.11, and 19.59, respectively, the average results of the time complexity for DSPL, LAPL and our method are  4.79, 689.01, and 0.48, respectively. {As seen,  our method is highly efficient compared with DSPL and LAPL in terms of both the memory and time costs}. 	
\begin{figure}[!t]
\centering
\includegraphics[width=\linewidth]{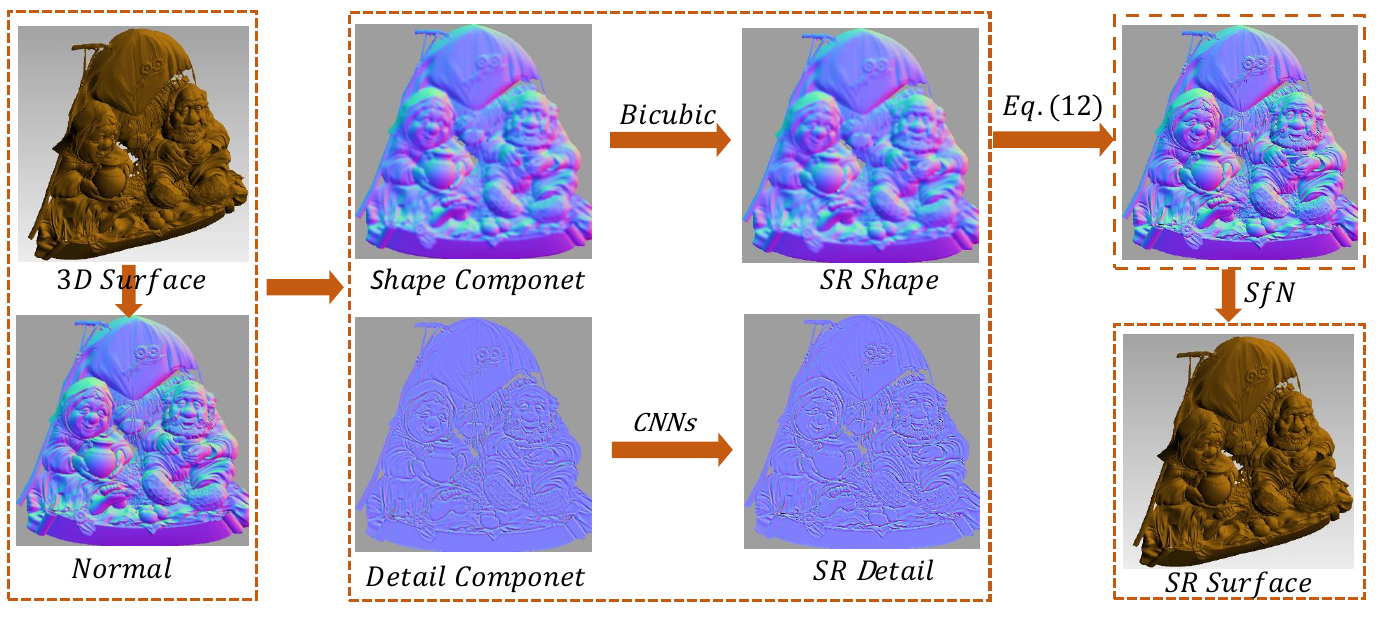}
\caption{\textbf{Pipeline of the proposed \textit{3D surface super-resolution}.} The \textit{shape component} and \textit{detail component} of the original normal are obtained by Eq. (\ref{eq_conv_cal}) and Eq. (\ref{eq_detail_form}), respectively. Then, the \textit{shape component} and the \textit{detail component} are up-sampled to the desired size through the bicubic interpolation and RDN-Net \cite{zhang2018residual}, respectively. Finally, the super-resolution result is obtained by Eq. (\ref{eq_transfer}) and reconstructed by the SfN method \cite{xie2014surface}.}
\label{fig_sr} 
\end{figure}

\begin{figure*}[!t]
\centering
\begin{tabular}{p{2.5cm}<{\centering} p{2.5cm}<{\centering} p{2.5cm}<{\centering} p{2.5cm}<{\centering} p{2.5cm}<{\centering} p{2.5cm}<{\centering}}
			\includegraphics[height=3.13cm]{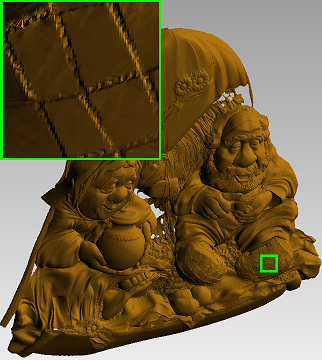}
			&\includegraphics[height=3.13cm]{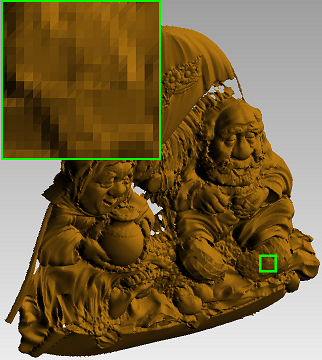}
			&\includegraphics[height=3.13cm]{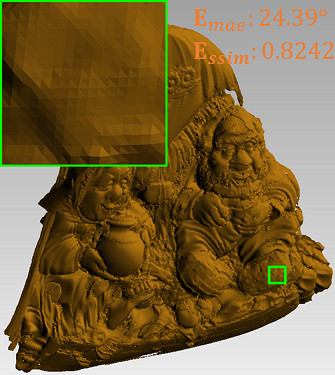}
			&\includegraphics[height=3.13cm]{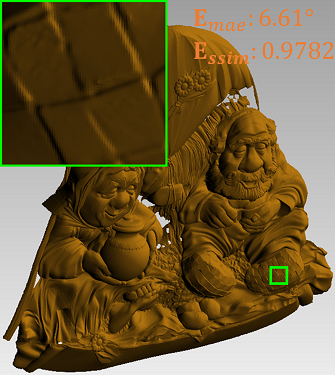}
			&\includegraphics[height=3.13cm]{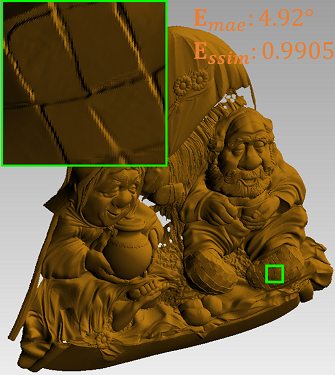}
            &\includegraphics[height=3.13cm]{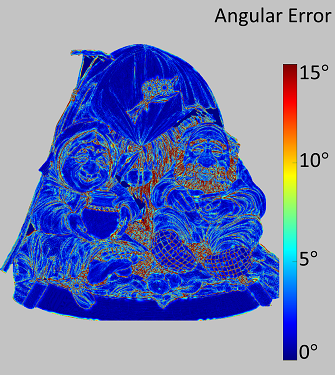}\\
			\footnotesize{\textit{Reference surface}} & \scriptsize{\textit{Low-resolution surface}} & \footnotesize{PU-Net \cite{yu2018pu}} &\footnotesize{RDN-Net  \cite{zhang2018residual}} & \footnotesize{\textit{Proposed}} &\footnotesize{{Energy map}}\\
\end{tabular} 
\caption{\textbf{Comparison results of \textit{3D surface super-resolution} for  the  \textit{Panno} model.} The $\mathbf{E}_{mae}$ and $\mathbf{E}_{ssim}$ results of the proposed method are $4.92^{\circ}$ and $0.9905$,  those of PU-Net \cite{yu2018pu} are $24.39^{\circ}$ and $0.8242$, and those of RDN-Net \cite{zhang2018residual} are $6.61^{\circ}$ and $0.9782$, respectively. {The angular error map of our method is shown on the most right-hand side}.} 
\label{fig_sr_pn} 
\end{figure*}

\begin{figure*}[!t]
	\centering
	\begin{tabular}{p{2.5cm}<{\centering} p{2.5cm}<{\centering} p{2.5cm}<{\centering} p{2.5cm}<{\centering} p{2.5cm}<{\centering} p{2.5cm}<{\centering}}
		\includegraphics[height=2.95cm]{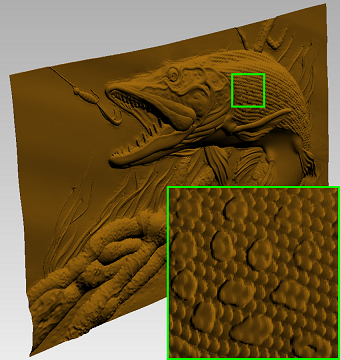}
		&\includegraphics[height=2.95cm]{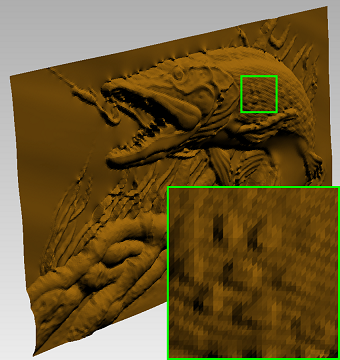}
		&\includegraphics[height=2.95cm]{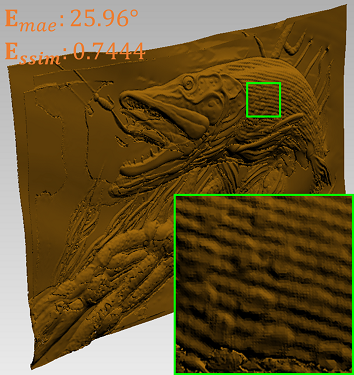}
		&\includegraphics[height=2.95cm]{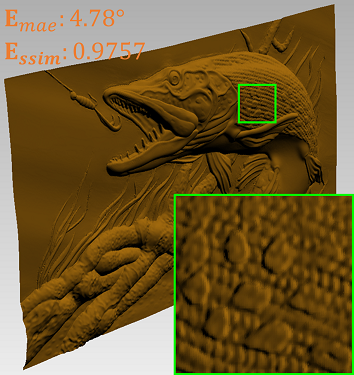}
		&\includegraphics[height=2.95cm]{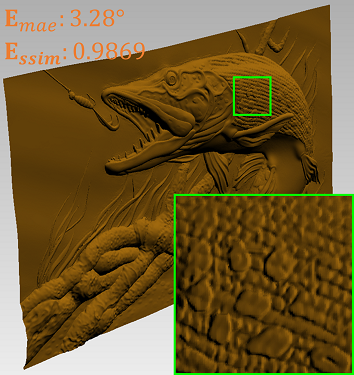}
        &\includegraphics[height=2.95cm,width=2.80cm]{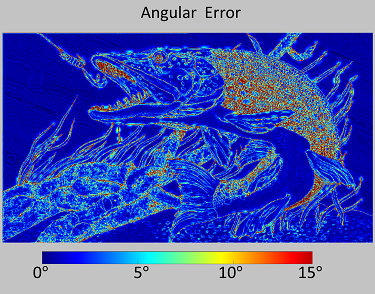}\\
		\footnotesize{\textit{Reference surface}} & \scriptsize{\textit{Low-resolution surface}} & \footnotesize{PU-Net \cite{yu2018pu}} &\footnotesize{RDN-Net  \cite{zhang2018residual}} & \footnotesize{\textit{Proposed}} &\footnotesize{{Energy map}}\\
	\end{tabular} 
	\caption{\textbf{Comparison results of \textit{3D surface super-resolution} for  the \textit{Fish} model.} The $\mathbf{E}_{mae}$ and $\mathbf{E}_{ssim}$ values of the proposed method are $3.28^{\circ}$ and $0.9869$,  those of PU-Net \cite{yu2018pu} are $25.96^{\circ}$ and $0.7444$, and those of RDN-Net \cite{zhang2018residual} are $4.78^{\circ}$ and $0.9757$, respectively. {The angular error map of our method is shown on the most right-hand side}.} 
	\label{fig_sr_fs} 
\end{figure*}

\begin{figure*}[!t]
\centering
\begin{tabular}{p{3cm}<{\centering} p{3cm}<{\centering} p{3cm}<{\centering} p{3cm}<{\centering} p{3cm}<{\centering} }
			\includegraphics[height=3cm]{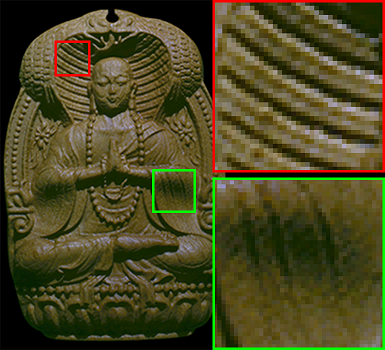}
			&\includegraphics[height=3cm]{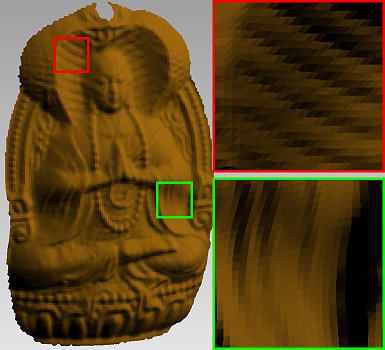}
			&\includegraphics[height=3cm]{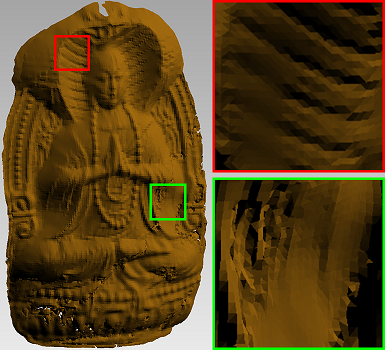}
			&\includegraphics[height=3cm]{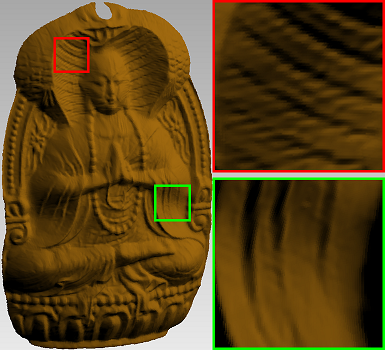}
			&\includegraphics[height=3cm]{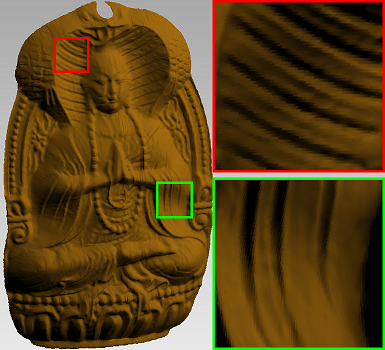}\\
			\small{\textit{Captured image}} & \small{\textit{Low-resolution surface}} & \small{PU-Net\cite{yu2018pu}} &\small{RDN-Net \cite{zhang2018residual}} & \small{\textit{Proposed}}\\
\end{tabular} 
\caption{\textbf{Comparison results of \textit{3D surface super-resolution} for  the  \textit{Woodcarving} model}. The associated normal map is obtained by our \textit{photometric stereo} device. Content in the red and green boxes is zoomed-in and shown on the right hand side for each image.}
\label{fig_sr_wc} 
\end{figure*}

\subsection{3D Surface Super-resolution}
The proposed \textit{detail component} can also be applied to 3D surface super-resolution based on existing learned CNN models. Specifically, for a low-resolution surface, we first decouple its normal map into the \textit{shape component} and the \textit{detail component}. Since the former is really smooth, it can be up-sampled by a simple method such as  \textit{Bicubic}. Since the latter is really complex, it can be enhanced by an advanced image-based super-resolution network. {The enhanced \textit{detail component} and \textit{shape component}} are then converted back, and finally the super-resolution 3D surface can be generated. The pipeline of our surface super-resolution scheme is illustrated in Fig. \ref{fig_sr}.

In principle, the proposed \textit{detail component} can be compatible with all existing image super-resolution networks if they are properly trained in the normal domain. To verify the performance, we have carried out three different comparison experiments to verify the super-resolution performance. Fig. \ref{fig_sr_pn} shows the original \textit{Panno} model with the size of 1812$\times$1800 downsampled by $\frac{1}{4} \times \frac{1}{4}$ first, and then used as the input for 4$\times$4  enhancement. PU-Net \cite{yu2018pu} provides a smooth result, the upsampled \textit{Panno} has the $\mathbf{E}_{mae}$ value of $24.39^\circ$ and the $\mathbf{E}_{ssim}$ value of $0.8242$. Compared with  PU-Net,  RDN-Net \cite{zhang2018residual} greatly restore the relevant surface details. Meanwhile, our method can effectively enhance the upsampled surface with the $\mathbf{E}_{mae}$ value of $4.92^\circ$ and the $\mathbf{E}_{ssim}$ value of $0.9905$, respectively.

Fig. \ref{fig_sr_fs} provides another 4$\times$4  \textit{Fish} super-resolution with the original size of 339$\times$192. As seen, the low-resolution surface has blurring fish scale. While PU-Net \cite{yu2018pu} can reduce the blurring effect, it can cause line-like artifact with the $\mathbf{E}_{mae}$ value of $25.96^\circ$ and the $\mathbf{E}_{ssim}$ value of $0.7444$. RDN-Net \cite{zhang2018residual} generates a better result than PU-Net, but the edge of fish scale is not obvious. In contrast, our proposed method achieves the best visual effect with the sharp edge of fish scale, where the $\mathbf{E}_{mae}$ value is  $3.28^\circ$ and the $\mathbf{E}_{ssim}$ value is  $0.9869$, respectively.

Fig. \ref{fig_sr_wc} illustrates a real-world low-resolution \textit{Woodcarving} model with the normal size of 126$\times$187 obtained by our \textit{photometric stereo} setup used for 4$\times$4 enhancement. In general, three methods can restore some surface details. However, when we zoom-in the relevant super-resolution results, our  method generates the best restoration performance compared with PU-Net \cite{yu2018pu} and RDN-Net \cite{zhang2018residual}.

\section{Discussions}
\label{sec:discussions}
In the proposed framework, a 3D surface is not expressed by meshes, voxels,  point clouds, or octrees \cite{wang2017cnn}. 
Instead, the orientation of a surface patch, known as the normal vector, is comprehensively explored as a new representation for 3D surface geometry details. 
To obtain the normal map of a surface, the surface points need to be projected onto the viewing plane to obtain the corresponding normal vector ``pixel''.  Our main focus is to investigate and validate the relationship between the shape and geometry details based on the single-view normal domain, and the quality can be affected if there are self-occlusions or discontinuities in an  input normal image.

{ As seen in Fig. \ref{fig_property_detail}, the error maps between the shape of ${{\bar \Delta }_{\mathcal{N}}}$ and $\mathbf{z}$ are dark-blue overall (with the average MAE value $0.86^\circ$), and large errors are shown in light-blue distributed over self-occluded surface regions, such as the nose boundary of \textit{Geothe} and the paw boundaries of \textit{Lizard}. The main reason is that normals on the self-occlusion regions are not properly captured in a single-view normal map, affecting the accuracy of the extracted \textit{detail component}.  The similar phenomenon can be observed in \textit{geometry detail transfer} (see Fig. \ref{fig_transfer_full}, Fig. \ref{fig_transfer_local_venus}, Fig. \ref{fig_transfer_local_bear}, and Fig. \ref{fig_transfer_local_buddha}) and \textit{3D surface super-resolution} (see Fig. \ref{fig_sr_pn} and Fig. \ref{fig_sr_fs}). 
It is worth noting that in addition to self-occlusion from the source surface, self-occlusion from the target surface also leads to major errors in \textit{geometry detail transfer}. A representative case can be found in Fig. \ref{fig_transfer_full}, where the target surface of  \textit{Bunny} contains a lot of self-occlusions (with the average MAE value $9.52^\circ$) on its neck, ears, tail, and gaps between legs. Fortunately, most of these local distortions are acceptable in our extensive experiments. For an interested surface with serious self-occlusion, multiple-view normal maps are required, which is discussed in the future work.}

In addition, the form of a normal map brings an extraordinary convenience to process 3D surface, which also means that it is not a ``what you see is what you get" paradigm. To obtain a 3D surface result, we have to resort to the SfN tool.  It is worth noting that the surface discontinuity is one of the main problems in the single-view reconstruction \cite{xie2019surface}, which often happens under the condition of self-occlusion due to the loss of depth information.
 This is also an open problem for all SfN methods. As a result, the final surface will also be affected by SfN in our proposed framework.

In summary, the detail theory we developed offers new insights into the 3D surface processing, enabling the two basic elements of 3D surface, shape and geometric details, to be {separately} expressed and effectively processed. This will bring the versatility of various image processing methods into 3D surface processing, which have been validated by some applications as described in Section \ref{sec:experiments}. 
Finally, while existing mesh-based methods show huge limitations in dense 3D surface processing due to their high computing time and excessive memory costs, our proposed has illustrated significant advantages in addressing these difficulties.

\section{Conclusion and Future Work}
\label{sec:conclusion}
In this article, a new surface detail representation is developed for 3D surface geometry processing in normal domain, where surface geometry details are separated from a surface normal map as a shape-uncorrelative feature with high transferability.  We also show that the dense laser-scanning point clouds can be satisfactorily processed when converting the orientation of each surface patch into a normal vector. To illustrate the versatility of our proposed, we further design three popular surface geometry detail processing algorithms based on the proposed normal-based framework. 
Experimental results validate the superiority of the proposed surface detail representation, and we believe that it will offer new insights into many 3D surface processing problems.

For future research, some further explorations can be developed in the following aspects to improve the performance of the proposed normal-based detail representation. First, the current design is simple and efficient for a single-view 3D surface, but their feature representation capabilities are limited to self-occlusion. A new normal map will be investigated to record both visible and invisible micro-geometry features in a single image, due to the fact that the current camera imaging projection, which is the main reason to result in occlusion features, is not a necessary way to record a surface normal. 
Second, the proposed method is effective on a single-view 3D surface. However, its performance on a multiple-view case \cite{li2020multi} is unknown and hence need to be investigated, where the self-occlusion or discontinuity may no longer be a problem. Third, due to the fact that the  quality of service for dynamic moving objects is becoming more and more popular, it is worthwhile to further extend this study to solving the geometric detail processing of dynamic point clouds (DPC).

\section{Acknowledgments}
The authors would like to give thanks to Mr. Maolin Cui for the preparations of the experimental data and the Python implementation, and would also like to express their sincere gratitude to the anonymous reviewers for valuable comments and suggestions.

{
\bibliographystyle{unsrt}
\bibliographystyle{IEEEtran}
\bibliography{sgdp_bib}
}

\end{document}